%% file: full.tex
\documentclass[journal,10pt,twocolumns,letter]{IEEEtran}

\usepackage{arabtex}
\usepackage{utf8}
\usepackage{booktabs}  
\usepackage[dvipsnames, table]{xcolor}
\usepackage{graphics}
\usepackage{bidipoem}

\usepackage{geometry}%
\usepackage[numbers,sort&compress]{natbib}

\usepackage[hyphens]{url}

\usepackage{subcaption}
\usepackage{mathtools}
\DeclarePairedDelimiter{\ceil}{\lceil}{\rceil}
\usepackage{amsfonts, amssymb, bm, amsmath}
\usepackage{caption}
\usepackage{verbatim}
\usepackage{enumitem}
\usepackage{graphicx, type1cm, lettrine}
\usepackage{qtree}
\usepackage{multirow}
\usepackage{soul}
\usepackage{collcell}
\usepackage{forest}
\usepackage{setspace}

\usepackage{fourier}
\DeclareMathAlphabet{\mathcal}{OT1}{pzc}{m}{it}
\DeclareSymbolFont{letters}{OML}{cmm}{m}{it}
\usepackage{tipa}
\definecolor{myBlue}{HTML}{7982db}

\definecolor{bgblue}{rgb}{0.41961,0.90784,0.90784}%
\definecolor{bgred}{rgb}{1,0.61569,0.61569}%
\definecolor{fgblue}{rgb}{0,0,0.6}%
\definecolor{fgred}{rgb}{0.6,0,0}%

\definecolor{flatBlue}{HTML}{B2B2FF}
\definecolor{flatRed}{HTML}{FFB2B2}
\definecolor{flatBrown}{HTML}{EAD8C5}
\definecolor{flatGrey}{HTML}{E1E3E3}
\definecolor{new_blue}{HTML}{517EB9}
\definecolor{hl}{HTML}{FFEF36}
\newcommand\gray{gray}

\newcommand\ColCell[1]{%
  \pgfmathparse{#1 <.8? 1: 0}%
  \ifnum\pgfmathresult=0\relax\color{white}\fi
\pgfmathparse{1-#1}%
\expandafter\cellcolor\expandafter[%
\expandafter\gray\expandafter]\expandafter{\pgfmathresult}#1}
\newcolumntype{E}{>{\collectcell\ColCell}c<{\endcollectcell}}

\usepackage{tikz}
\usepackage{pgf}
\usepackage{pgfplots}
\tikzset{
  Above/.style={
    midway,
    above,
    font=\scriptsize,
    text width=1.5cm,
    align=center,
  },
  Below/.style={
    midway,
    below,
    font=\scriptsize,
    text width=1.5cm,
    align=center
  }
}
\usetikzlibrary{arrows} 

\pgfplotsset{compat=1.11,
  /pgfplots/ybar legend/.style={
    /pgfplots/legend image code/.code={
      \draw[##1,/tikz/.cd,bar width=1pt,yshift=0.0cm,bar shift=0pt]
      plot coordinates {(0cm,0.0cm)};},
  },
}

\usepackage[font=footnotesize]{caption}

\input{pgfplot_configurations.tex}

\begin{document}
\setcode{utf8}

\title{Learning meters of Arabic and English poems with Recurrent Neural Networks: a step forward
  for language understanding and synthesis}

\author{%
  Waleed~A.~Yousef\textsuperscript{a},~\IEEEmembership{Senior Member,~IEEE};~\thanks{Waleed
    A. Yousef is an associate professor, \protect\url{wyousef@fci.helwan.edu.eg}}

  Omar M. Ibrahime\textsuperscript{a,b};~\thanks{Omar M. Ibrahime, B.Sc., \protect\url{umar.ibrahime@fci.helwan.edu.eg}}
  Taha M. Madbouly\textsuperscript{a,b};~\thanks{Taha M. Madbouly, B.Sc., \protect\url{tahamagdy@fci.helwan.edu.eg}} 
  Moustafa A. Mahmoud\textsuperscript{a,b};~\thanks{Moustafa A. Mahmoud, B.Sc., Senior Big Data Engineer, \protect\url{mustafa.mahmoud@fci.helwan.edu.eg}}


  \thanks{\textsuperscript{a}Human Computer Interaction Laboratory (HCILAB:\
    \protect\url{www.hciegypt.com}); and Department of Computer Science, Faculty of Computers and Information,
    Helwan University, Egypt.}
  \thanks{\textsuperscript{b}These three authors contributed equally
    to the manuscript, their names are ordered alphabetically according to the family name, and each
    of them is the second author.}
}

\maketitle

\begin{abstract}
  Recognizing a piece of writing as a poem or prose is usually easy for the majority of people; however,
  only specialists can determine which meter a poem belongs to. In this paper, we build Recurrent
  Neural Network (RNN) models that can classify poems according to their meters from plain text. The
  input text is encoded at the character level and directly fed to the models without feature
  handcrafting. This is a step forward for machine understanding and synthesis of languages in
  general, and Arabic language in particular.

  Among the 16 poem meters of Arabic and the 4 meters of English the networks were able to correctly
  classify poem with an overall accuracy of 96.38\% and 82.31\% respectively. The poem datasets used to
  conduct this research were massive, over 1.5 million of verses, and were crawled from different
  nontechnical sources, almost Arabic and English literature sites, and in different heterogeneous
  and unstructured formats. These datasets are now made publicly available in clean, structured, and
  documented format for other future research.

  To the best of the authors' knowledge, this research is the first to address classifying poem
  meters in a machine learning approach, in general, and in RNN featureless based approach, in
  particular. In addition, the dataset is the first publicly available dataset ready for the purpose
  of future computational research.
\end{abstract}

\begin{IEEEkeywords}
  Poetry, Meters, Al-'arud, Arabic, English, Recurrent Neural Networks, RNN, Deep Learning, Deep Neural
  Networks, DNN, Classification, Text Mining.
\end{IEEEkeywords}

\section{Introduction}\label{sec:introduction}

\subsection{Arabic Language}\label{sec:arabic-language}
Arabic is the fifth most widely spoken language~\cite{Simons201720thEditionEthnologue}.  It is
written from right to left (RTL). Its alphabet consists of 28 primary letters and 8 further derived
letters from the primary ones, which makes all letters sum up to 36.  The writing system is cursive;
hence, most letters are joined and a few letters remain disjoint.

\begin{table}[!t]
  \centering
  \resizebox{\columnwidth}{!}{%
    \begin{tabular}{c c c c c c}
      \toprule
      \textbf{\small{Diacritics}}  & {\small{\textit{without}}} & {\small{\textit{fat-ha}}} & {\small{\textit{dam-ma}}} & {\small{\textit{kas-ra}}} & {\small{\textit{sukun}}} \\
      \midrule
      \textbf{\small{writing}}     & \<د>             & \<دَ>            & \<دُ>            & \<دِ>            & \<دْ>\\
      \textbf{\small{short vowel}} & ---                         & /\textit{a}/              & /\textit{u}/              & /\textit{i}/              & /\textit{no vowel}/\\
      \bottomrule
    \end{tabular}
  }
  \caption{\textit{The 4 Diacritics of Arabic Language. Transliterated names (1st row), writing
      style on example letter \<د>} (2nd row), and corresponding short pronunciation
    vowel (3rd row).}\label{arabic:diacritics_dal}
\end{table}

Each Arabic letter represents a consonant, which means that short vowels are not represented by the
36 letters. For this reason the need rises for \textit{diacritics}, which are symbols ``decorating''
original letters. Usually, a \textit{diacritic} is written above or under the letter to emphasize the
short vowel accompanied with that letter. There are 4 diacritics: \mbox{\<◌َ> \<◌ُ>
  \<◌ِ> \<◌ْ>}. Table~\ref{arabic:diacritics_dal} lists these 4 diacritics on an
example letter \<د>, their transliterated names, along with their short vowel
representation. Each of the three diacritics \mbox{\<◌َ> \<◌ُ> \<◌ِ>} is called \textit{harakah}; whereas the fourth \<◌ْ> is called
\textit{sukun}. Diacritics are just to make short vowels clearer; however, their writing is not
compulsory since they can be almost inferred from the grammatical rules and the semantic of the
text. Moreover, a phrase with diacritics written for only some letters is linguistically sound.

There are two more sub-diacritics made up of the basic four. The first is known as \textit{shaddah}
\<◌ّ>, which must associate with one of the three \textit{harakah} and written as
\mbox{\<◌َّ> \<◌ُّ> \<◌ِّ>}. \textit{Shaddah} is a shorthand writing for the
case when a letter appears two times in a row where the first occurrence is accompanied with
\textit{sukun} and the second occurrence is accompanied with \textit{harakah}. Then, for short, it
is written as one occurrence accompanied with \textit{shaddah} associated with the corresponding
\textit{harakah}. E.g., \mbox{\<دْدَ>} is written as \<دَّ>. The second is known as
\textit{tanween}, which must associate as well one of the three \textit{harakah} and written as:
\mbox{\<◌ٍ> \<◌ٌ> \<◌ً>}. \textit{Tanween} accompanies the last letter of
some words, according to Arabic grammar, ending with \textit{harakah}. This is merely for reminding
the reader to pronounce the word as if there is \<نْ> (sounding as \textit{/n/}), follows
that \textit{harakah}. However, it is just a phone and is not a part of the word itself. E.g.,
\<رَجُلٌ> is pronounced \mbox{\<رَجُلُ + نْ>} and \<رَجُلٍ> is pronounced
\mbox{\<رَجُلِ + نْ>}.

\subsection{Arabic Poetry (\<الشعر العربى>)}\label{sec:arab-poetry-text}
Arabic poetry is the earliest form of Arabic literature; it dates back to the sixth century. Poets
wrote poems without knowing exactly what rules make a collection of words a poem. People recognize
poetry by nature, but only talented ones who could write poems. This was the case until
\textit{Al-Farahidi} (718 – 786 CE) has analyzed Arabic poems and recognized their patterns. He came
up with that the succession of consonants and vowels, and hence \textit{harakah} and \textit{sukun},
rather than the succession of letters themselves, produces patterns (\textit{meters}) which keeps
the balanced music of pieces of poem. He recognized fifteen meters. Later, one of his students,
\textit{Al-khfash}, discovered one more meter to make them all sixteen. Arabs call meters
\<بحور>, which means ``seas''~\cite{Moustafa}.

A poem is a collection of verses. A verse example is:%

\begin{traditionalpoem*}
\<قفا نبك من ذِكرى حبيب ومنزل>
\quad & \quad
\<بسِقطِ اللِّوى بينَ الدَّخول فحَوْملِ>
\end{traditionalpoem*}

A verse, known in Arabic as \textit{bayt} \<(بَيت)>, which consists of two halves. Each
half is called a \textit{shatr} (\<شطر>).  \textit{Al-Farahidi} has introduced
\textit{al-'arud} (\<العروض>), which is often called the \textit{Knowledge of Poetry} or
the study of poetic meters. He laid down rigorous rules and measures, with them we can determine
whether a meter of a poem is sound or broken. For the present article to be fairly self-contained,
where many details are reduced, a very brief introduction to \textit{al-'arud} is provided through
the following lines.
\begin{table}[!tb]
  \centering
  \resizebox{\columnwidth}{!}{%
    \begin{tabular}[t!]{ccccccccc}
      \toprule
      \textbf{Foot}& \<فَعُوْلُنْ> & \<فَاْعِلُنْ> & \<مُسْتَفْعِلُنْ> & \<مَفَاْعِيْلُنْ> &\<مَفْعُوْلَاْتُ> &\<فَاْعِلَاْتُنْ> &\<مُفَاْعَلَتُنْ> &\<مُتَفَاْعِلُنْ>\\
      \midrule
      \textbf{Scansion}&\texttt{0/0//}&\texttt{0//0/}&\texttt{0//0/0/}&\texttt{0/0/0//}&\texttt{/0/0/0/}&\texttt{0/0//0/}&\texttt{0///0//}&\texttt{0//0///}\\
      \bottomrule
    \end{tabular}}
  \caption{The eight feet of Arabic poetry. Every digit (\texttt{/} or \texttt{0}) represents the
    corresponding diacritic over a letter of a feet: \textit{harakah} (\mbox{\<◌َ>
      \<◌ُ> \<◌ِ>}) or \textit{sukun} (\<◌ْ>) respectively. Any of the
    three letters \mbox{\<و ا ى>} (called \textit{mad}) is equivalent to \texttt{0};
    \textit{tanween} and \textit{shaddah} are equivalent to \texttt{0/} and \texttt{/0}
    respectively.}\label{arud:feet}
\end{table}

A meter is an ordered sequence of phonetic syllables (blocks or mnemonics) called \textit{feet}. A
foot is written with letters only having \textit{harakah} or \textit{sukun}, i.e., with neither
\textit{shaddah} nor \textit{tanween}; and hence each letter in a foot maps directly to either a
consonant or a vowel. Therefore, feet represent phonetic mnemonics, of the pronounced
poem, called \textit{tafa'il} (\<تفاعيل>). Table~\ref{arud:feet} lists the eight feet used
by Arabs and the pattern (scansion) of \textit{harakah} and \textit{sukun} of each foot, where a
\textit{harakah} is represented as \texttt{/} and a \textit{sukun} is represented as
\texttt{0}. Each scansion reads RTL to match the letters of the corresponding foot.

According to \textit{Al-Farahidi} and his student, they discovered sixteen combinations of
\textit{tafa'il} in Arabic poems; they called each combination a \textit{meter}
(\<بحر>). (Theoretically speaking, there is no limit for either the number of
\textit{tafa'il} or their combinations; however, Arab composed poems using only this structure). A
meter appears in a \textit{verse} twice, once in each \textit{shatr}. E.g., 
\mbox{\<وَيُسْأَلُ فِي الحَوادِثِ ذو صَواب>} is the first \textit{shatr} of a verse of \textit{Al-Wafeer} meter
\mbox{\<مُفَاْعَلَتُنْ مُفَاْعَلَتُنْ فَعُوْلُنْ>}. The pattern of the \textit{harakah} and \textit{sukun} of
this meter is \mbox{\texttt{0/0// 0///0// 0///0//}} (RTL), and is obtainable directly by replacing
each of the three feet by the corresponding code in table~\ref{arud:feet}. This pattern corresponds
exactly to the pattern of \textit{harakah} and \textit{sukun} of the pronounced (not written)
\textit{shatr}. E.g., the pronunciation of the first two words and the first two letters of the
third word \mbox{\textcolor{fgred}{\<فِى الْ>} \textcolor{OliveGreen}{\<وَيُسْاَلُ>}} has exactly the same
pattern as the first of the three \textit{{tafa'il}} of the meter
\mbox{\textcolor{fgred}{\<تُنْ>}\textcolor{OliveGreen}{\<مُفَاعَلَـ>}}, and both have the scansion
\texttt{\color{fgred}{0/}\color{OliveGreen}{//0//}}. For more clarification, the colored parts have
corresponding pronunciation pattern; which emphasizes that the start and end of a word do not have
to coincide with the start and end of the phonetic syllable. The pronunciation of the rest of the
\textit{shatr} \mbox{\<حوادث ذو صواب>} maps to the rest of the meter
\mbox{\<مفاعلتن فعولن>}. Any other poem, regardless of its wording and semantic, following
the same meter, i.e., following the same pattern of \textit{harakah} and \textit{sukun}, will have
the same pronunciation or phonetic pattern.
\begin{table}[!tb]
  \centering
  \resizebox{\columnwidth}{!}{%
    \begin{tabular}[t!]{lrr}
      \toprule
      \textbf{Meter Name}   & \textbf{Pattern} & \textbf{Scansion}\\
      \midrule
      \textit{al-Taweel}    & \<فَعُوْلُنْ مَفَاْعِيْلُنْ فَعُوْلُنْ مَفَاْعِلُنْ> & \texttt{0//0// ~~0/0// 0/0/0// ~~0/0//}  \\
      \textit{al-Kamel}     & \<مُتَفَاْعِلُنْ مُتَفَاْعِلُنْ مُتَفَاْعِلُنْ>     & \texttt{0//0/// 0//0/// 0//0///}\\
      \textit{al-Baseet}    & \<مُسْتَفْعِلُنْ فَاْعِلُنْ مُسْتَفْعِلُنْ فَاْعِلُنْ> & \texttt{0//0/ 0//0/0/ ~~0//0/ 0//0/0/} \\
      \textit{al-Khafeef}   & \<فَاْعِلَاْتُنْ مُسْتَفْعِلُنْ فَاْعِلَاْتُنْ>     & \texttt{0/0//0/ 0//0/0/ 0/0//0/}\\
      \textit{al-Wafeer}    & \<مُفَاْعَلَتُنْ مُفَاْعَلَتُنْ فَعُوْلُنْ>      & \texttt{0/0// 0///0// 0///0//}\\
      \textit{al-Rigz}      & \<مُسْتَفْعِلُنْ مُسْتَفْعِلُنْ مُسْتَفْعِلُنْ>     & \texttt{0//0/0/ 0//0/0/ 0//0/0/}\\
      \textit{al-Raml}      & \<فَاْعِلَاْتُنْ فَاْعِلَاْتُنْ فَاْعِلَاْتُنْ>     & \texttt{0/0//0/ 0/0//0/ 0/0//0/}\\
      \textit{al-Motakarib} & \<فَعُوْلُنْ فَعُوْلُنْ فَعُوْلُنْ فَعُوْلُنْ>     & \texttt{0/0// ~~0/0// ~~0/0// ~~0/0//}\\
      \textit{al-Sar'e}     & \<مُسْتَفْعِلُنْ مُسْتَفْعِلُنْ مَفْعُوْلَاْتُ>     & \texttt{/0/0/0/ 0//0/0/ 0//0/0/}\\
      \textit{al-Monsareh}  & \<مُسْتَفْعِلُنْ  مَفْعُوْلَاْتُ مُسْتَفْعِلُنْ>    & \texttt{0//0/0/ /0/0/0/ 0//0/0/}\\
      \textit{al-Mogtath}   & \<مُسْتَفْعِلُنْ فَاْعِلَاْتُنْ  فَاْعِلَاْتُنْ>    & \texttt{0/0//0/ 0/0//0/ 0//0/0/}\\
      \textit{al-Madeed}    & \<فَاْعِلَاْتُنْ فَاْعِلُنْ فَاْعِلَاْتُنْ>       & \texttt{0/0//0/ ~~0//0/ 0/0//0/}\\
      \textit{al-Hazg}      & \<مَفَاْعِيْلُنْ مَفَاْعِيْلُنْ>             & \texttt{0/0/0// 0/0/0//}\\
      \textit{al-Motadarik} & \<فَاْعِلُنْ فَاْعِلُنْ فَاْعِلُنْ فَاْعِلُنْ>     & \texttt{0//0/ ~~0//0/ ~~0//0/ ~~0//0/}\\
      \textit{al-Moktadib}  & \<مَفْعُوْلَاْتُ مُسْتَفْعِلُنْ مُسْتَفْعِلُنْ>     & \texttt{0//0/0/ 0//0/0/ /0/0/0/}\\
      \textit{al-Modar'e}   & \<مَفَاْعِيْلُنْ فَاْعِلَاْتُنْ فَاْعِلَاْتُنْ>     & \texttt{0/0//0/ 0/0//0/ 0/0/0//}\\
      \bottomrule
    \end{tabular}}
  \caption{The sixteen meters of Arabic poem: transliterated names (1st col.), mnemonics or
    \textit{tafa'il} (2nd col.), and the corresponding pattern of \textit{harakah} and
    \textit{sukun} in \texttt{0/} format or scansion (3rd col.).}\label{arud:meters}
\end{table}

Table~\ref{arud:meters} lists the names of all the sixteen meters, the transliteration of their
names, and their patterns (scansion). Each pattern is written in two equivalent forms: the feet
style using the eight feet of Table~\ref{arud:feet} and the scansion pattern using the \texttt{0/}
symbols. The scansion is written in groups; each corresponds to one foot and all are RTL\@.

\subsection{English poetry}\label{sec:english-poetry}
English poetry dates back to the seventh century. At that time poems were written in
\textit{Anglo-Saxon}, also known as \textit{Old English}. Many political changes have influenced
the language until it became as it is nowadays. English prosody was not formalized rigorously as a
stand-alone knowledge, but many tools of the \textit{Greek} prosody were borrowed to describe it.

A \textit{syllable} is the unit of pronunciation having one vowel, with or without surrounding
consonants. English words consist of one or more syllables. For example the word \mbox{``water''}
(pronounced as \mbox{\textipa{\sffamily /"wO:t@(r)/}}) consists of two phonetic syllables:
\mbox{\textipa{\sffamily /"wO:/}} and \mbox{\textipa{\sffamily /t@(r)/}}. Each syllable has only one
vowel sound. Syllables can be either stressed or unstressed and will be denoted by \textit{/} and
$\times$ respectively. In phonology, a stress is a phonetic emphasis given to a syllable, which can
be caused by, e.g., increasing the loudness, stretching vowel length, or changing the sound
pitch. In the previous ``water'' example, the first syllable is stressed, which means it is
pronounced with high sound pitch; whereas the second syllable is unstressed which means it is
pronounced in low sound pitch. Therefore, ``water'' is a stressed-unstressed word, which can be
denoted by \mbox{\textit{/}$\times$}. Stresses are shown in the phonetic script using the primary stress
symbol \textipa{\sffamily (")}.
\begin{table}[!tb]
  \centering
  \resizebox{\columnwidth}{!}{%
    \begin{tabular}{cccccccc}
      \toprule
      \textbf{{Foot}} & \textit{Iamb} & \textit{Trochee} & \textit{Dactyl} & \textit{Anapest} & \textit{Pyrrhic} & \textit{Amphibrach} & \textit{Spondee} \\
      \midrule
      \textbf{Stresses}& $\times$\textit{/} & \textit{/}$\times$ & \textit{/}$\times\times$ & $\times\times$\textit{/} & $\times\times$ & $\times$\textit{/}$\times$ & \textit{/}\textit{/} \\
      \bottomrule
    \end{tabular}}
  \caption{The seven feet of English poem. Every foot is a combination of stressed and unstressed
    syllables, denoted by \textit{/} and \textit{x} respectively.}\label{feet}
\end{table}
There are seven different combinations of stressed and unstressed syllables that make the seven
poetic \textit{feet}.  They are shown in table~\ref{feet}. Meters are described as a sequence of
feet. English meters are \textit{qualitative} meters, which are stressed syllables coming at regular
intervals. A meter is defined as the repetition of one of the previous seven feet one to eight
times. If the foot is repeated once, then the verse is \textit{monometer}, if it is repeated twice
then it is a \textit{dimeter} verse, and so on until \textit{octameter} which means a foot is
repeated eight times. This is an example, where stressed syllables are bold: \mbox{``That
  \textbf{time} of \textbf{year} thou \textbf{mayst} in \textbf{me} be\textbf{hold}''}. The previous
verse belongs to the \textit{Iamb} meter, with the pattern \mbox{$\times$\textit{/}} repeated five times;
so it is an \textit{Iambic pentameter} verse.

\subsection{Paper Organization}\label{sec:paper-organization}
The rest of this paper is organized as follows. Sec.~\ref{sec:literature-review} is a literature
review of meter detection of both languages; the novelty of our approach and the point of departure
from the literature will be emphasized. Sec.~\ref{sec:datasets} explains the data acquisition steps
and the data repository created by this project to be publicly available for future research; in
addition, this section explains character encoding methods, along with our new encoding method and
how they are applied to Arabic letters in particular. Sec.~\ref{sec:model} explains how experiments
are designed and conducted in this research. Sec.~\ref{sec:results} presents and interprets the
results of these experiments. Sec.~\ref{sec:discussion} is a discussion, where we emphasize the
interpretation of some counter-intuitive results and connect them to the size of conducted
experiments, and the remedy in the future work that is currently under implementation.

\section{Literature review}\label{sec:literature-review}
To the best of our knowledge, the problem addressed in the present paper has never been addressed in
the literature. ``Learning'' poem style from text so that machines are able to classify unseen
written poem to the right meter seems to be a novel area. However, there is some literature on
recognizing the meters of written Arabic poem using rule-based deterministic algorithms. We did not
find related work on English written poem. These rules are derived by humans/experts and not learned
by machines from data. In this regard, this is quite irrelevant to our present problem, and this is
our point of departure in this research. However, we review these methods for the sake of
completion.

\cite{Kurt2012AlgorithmForDetectionAnalysis} worked on the Ottoman Language. They converted the
Ottoman text into a lingual form; in particular, the poem was transliterated to Latin transcription
alphabet (LTA). Next, the text was fed to the algorithm, which uses a database containing all
Ottoman meters, to be compared to the existing meters and then classified to the closest one.

\cite{Alnagdawi2013FindingArabicPoemMeter} worked on Arabic language. They formalized the
\textit{scansion}s, \textit{al-'arud}, and some lingual rules (like pronounced and silent rules,
which are directly related to \textit{harakah} and \textit{sukun}) in terms of context-free grammar
and regular expression templates. The classification accuracy was only 75\% on a very small sample
of 128 verses.

\cite{Abuata2016RuleBasedAlgorithmFor} worked on Arabic language. They designed a five-step
deterministic \textit{algorithm} for analyzing and detecting meters. First, they input text carrying
full diacritics for all letters. Second, they convert the input text into \textit{al-'arud} writing
style (Sec.~\ref{sec:arab-poetry-text}) using \texttt{if-else} rules. Third, the metrical
\textit{scansion} rules are applied, which leaves the input text as a sequence of zeros and
ones. Fourth, each group of zeros and ones are defined as a \textit{tafa'il}
(Table~\ref{arud:feet}). Finally, the input text is classified to the closest meter to the
\textit{tafa'il} sequence (Table~\ref{arud:meters}). The classification accuracy of this algorithm
is 82.2\%, on a relatively small sample of 417 verses.

\bigskip

It is quite important to observe that although these algorithms are deterministic rules that are fed
by experts, alas, they did not succeed in producing high accuracy, 75\% and 82.2\%. This is in
contrast to our featureless RNN approach that remarkably outperforms these methods by achieving
96.38\%. The interpretation of that is clear. The rule-based algorithms cannot list all possible
combinations of anomalies in written text, including missing diacritics on some characters, breaking
the meter by a poet, etc; whereas, RNN will be able to ``learn'' by example the probability of these
occurrences. Table~\ref{tab:summ-results} summarizes the accuracies and the testing sample size of
this literature in comparison with our approach. It is even more surprising that while these
algorithms must work on poem with diacritics, RNN accuracy only dropped about 1\% when trained on
plain poem with no diacritics.

\begin{table}[!tb]
  \centering
  \begin{tabular}{c c c c}
    \toprule
    \textbf{Ref.}& \textbf{Accuracy}& \textbf{Test Size} & \textbf{Poem}\\
    \midrule
    \cite{Alnagdawi2013FindingArabicPoemMeter}   & 75\%     & 128     & \multirow{3}{*}{Arabic}\\
    \cite{Abuata2016RuleBasedAlgorithmFor}      & 82.2\%   & 417     & \\
    This article   & 96.38\%  & 150,000 & \\
    \midrule
    This article   & 82.31\%  &  1,740  & English\\
    \bottomrule
  \end{tabular}
  \caption{Overall accuracy of this article compared to literature.}\label{tab:summ-results}
\end{table}

\section{Datasets: Acquisition, Encoding, and Repository}\label{sec:datasets}
Sec.~\ref{sec:arabic_english_dataset} explains how the Arabic and English datasets were scraped from
different non-technical web sources; and hence needed a lot of cleaning and structuring. For future research on
these datasets, and probably for collecting more poem datasets, we launched the data repository
``Poem Comprehensive Dataset (PCD)''~\citep{Yousef2018PoemComprehensiveDataset} that is publicly
available for the whole community. The datasets on this repository are in their final clean formats
and ready for computational purposes. Sec.~\ref{sec:data-encoding} explains the data encoding at the
character level before feeding to the RNN\@.

\subsection{Arabic and English Datasets Acquisition}\label{sec:arabic_english_dataset}
\begin{figure}[!tb]
  \centering
  \begin{tikzpicture}[scale=0.8]
    \input{dataset_size_ar.tex}
    \input{dataset_size_en.tex}
  \end{tikzpicture}%
  \caption{\footnotesize Class size (number of verses), of both Arabic and English datasets, ordered
    descendingly on $y$ axis vs.\ corresponding meter name on $x$
    axis.}\label{fig:footn-footn-class}
\end{figure}
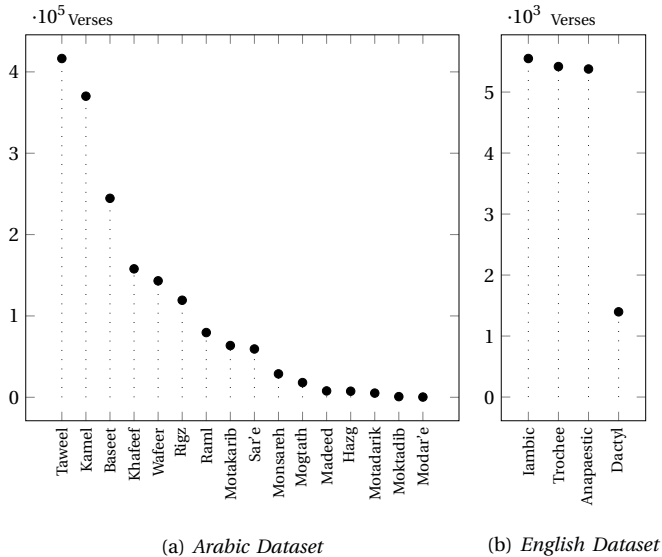

We have scrapped the Arabic dataset from two big poetry websites~\citep{diwan,
  PoetryEncyclopedia2016}; then both are merged into one large dataset. The total number of verses
is 1,862,046. Each verse is labeled by its meter, the poet who authored it, and the age it belongs
to. Overall, there are 22 meters, 3701 poets and 11 ages: Pre-Islamic, Islamic, Umayyad, Mamluk,
Abbasid, Ayyubid, Ottoman, Andalusian, the era between Umayyad and Abbasid, Fatimid, and modern. We
are only interested in the 16 classic meters which are attributed to \textit{Al-Farahidi}. These
meters comprise the majority of the dataset with a total number of 1,722,321
verses. Figure~\ref{fig:footn-footn-class}-a is an ordered bar chart of the number of verses per
meter. It is important to mention that the state of verse diacritic is inconsistent; a verse can
carry full, partial, or no diacritics. This should affect the accuracy results as discussed in
Sec.~\ref{sec:discussion}.

The English dataset is scraped from many different web
resources~\cite{HuberEighteenthCenturyPoetryArchive}. It consists of 199,002 verses; each of them is
labeled with one of the four meters: \textit{Iambic}, \textit{Trochee}, \textit{Dactyl} and,
\textit{Anapaestic}. Since the \textit{Iambic} class dominates the dataset with 186,809 verses, we
downsampled it to 5550 verses to keep classes almost balanced. Figure~\ref{fig:footn-footn-class}-b
is an ordered bar chart of the number of verses per meter.

\bigskip

For both datasets, data cleaning was tedious but necessary step
before direct computational use. The poem contained non-alphabetical characters, unnecessary in-text
white spaces, redundant glyphs, and inconsistent diacritics. E.g., the Arabic dataset in many places
contained two consecutive \textit{harakah} on the same letter or a \textit{harakah} after a white
space. In addition, as a pre-encoding step, we have factored a letter having either \textit{shaddah}
or \textit{tanween} into two letters, as explained in Sec.~\ref{sec:arabic-language}. This step
shortens the encoding vector and saves more memory as explained in the next section. Each of the
Arabic and English datasets, after merging and cleaning, is labeled and structured in its final
format that is made publicly available \citep{Yousef2018PoemComprehensiveDataset} as introduced
above.

\subsection{Data Encoding}\label{sec:data-encoding}
\begin{figure*}[!tb]
  \centering
  \input{encoding_three_figures_together.tex}
  \caption{Three encoding schemes: \textit{One-hot} (a), \textit{binary} (b), and \textit{two-hot}
    (c). The example word \<مَرْحَبَا> consists of 5 letters and is used to illustrate the
    \textit{one-hot} and \textit{binary} encodings. One of its letters \<بَ> is selected as
    an example to illustrate the \textit{two-hot} encoding (c).}\label{fig:One-Binary-Encoding}
\end{figure*}
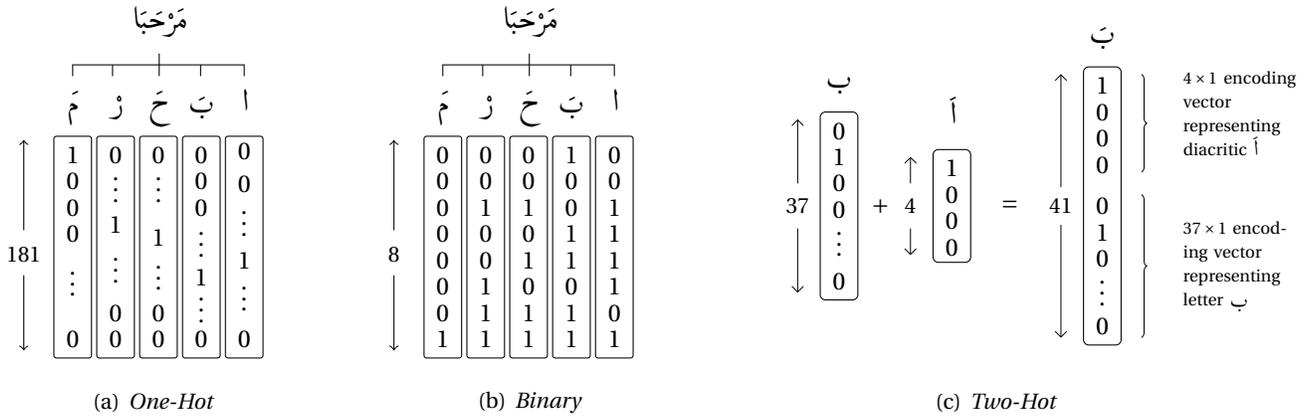
It was introduced in Sec.~\ref{sec:arab-poetry-text} that a poem meter, in particular Arabic poem,
is a phonetic pattern of vowels and consonants that is inferred from \textit{harakah} and
\textit{sukun} of the written text. It is therefore obvious that text should be fed to the network
at the character (not word) level. Characters are categorical predictors, and therefore character
encoding is necessary for feeding them to any form of Neural Networks (NN). Categorical variable
encoding has an impact on the neural network performance. (We elaborate on that upon discussing the
results in Sec.~\ref{sec:discussion}). E.g.,~\cite{Potdar2017ComparativeStudyCategoricalVariable} is
a comparative study for six encoding techniques. They have trained NN on the \textit{car evaluation}
dataset after encoding the seven ordered qualitative
features.~\cite{Agirrezabal2017ComparisonFeatureBasedNeural} shows that representations of data
learned from character-based neural models are more informative than the ones from hand-crafted
features.

In this research, we have used the two known encoding schemes \textit{one-hot} and \textit{binary},
in addition to the \textit{two-hot} that we introduced for more efficient encoding of the Arabic
letters. Before explaining these three encoding schemes, we need to make the distinction clear
among: letters, diacritics, characters (or symbols), and encoding vectors. In English language (and
even in Latin that has letters with diacritics, e.g., \textit{ê, é, è, ë, ē, ĕ, ě}), each letter is
considered a standalone character (or symbol) with a unique Unicode. Each of them is encoded to a
vector, whose length $n$ depends on the encoding scheme. Then, a word, or a verse, consisting of $p$
letters (or characters in this case) would be represented as $n \times p$ matrix. However, in Arabic
Language, diacritics are treated differently in the Unicode system. A diacritic is considered a
standalone character (symbol) with a unique Unicode (in contrast to Latin diacritics as just
explained). E.g., the Arabic letter \<بَ>, which is the letter \<ب> accompanied
with the diacritic \<◌َ>, is considered in Unicode system as two consecutive characters,
the character \<ب> followed by the character \<◌َ>, where each has its own
Unicode. Based on that, Arabic and English text are encoded using each of the three encoding methods
as follows.

\bigskip

\subsubsection{One-Hot encoding}\label{sec:one-hot-encoding}
In English, there are 26 letters, a white-space, and an apostrophe; hence, there are 28 final
characters. In \textit{one-hot} encoding each of the 28 characters will be represented by a vector
of length $n=28$ having a single one and 27 zeros; hence, this is a sparse encoding. In Arabic, we
will represent a combination of a letter and its diacritic together as a single encoding
vector. Since, from Sec.~\ref{sec:arabic-language}, there are 36 letters, 4 diacritics and a
white-space, and since a letter may or may not have a diacritic whereas the white-space cannot,
there is a total of $36 \times (4+1) + 1 = 181$ combinations. Hence, the encoding vector length is
$n=181$; each vector will have just a single one and 180 zeros.

\subsubsection{Binary Encoding}\label{sec:binary-encoding}
In binary encoding, an encoding vector of length $n$ contains a unique binary combination in
contrast to the sparse \textit{one-hot} encoding representation. Therefore, the encoding lengths of
English and Arabic are $\ceil{\log_2 28} = 5$ and $\ceil{\log_2 181} = 8$ respectively, which is a
huge reduction in dimensionality. However, this will be on the expense the challenge added to find
the best network architecture design that is capable of decoding this scheme
(Sec.~\ref{sec:discussion}).

\subsubsection{Two-Hot encoding}\label{sec:two-hot-encoding}
For Arabic language, where diacritics explode the length of the \textit{one-hot} encoding vector to
181, we introduce this new encoding. In this encoding, the 36 letters and the white-space on a hand
and the 4 diacritics on the other hand are encoded separately using two \textit{one-hot} encoding
vectors of lengths $n=37$ and $n=4$ respectively. The final \textit{two-hot} encoding of a
letter with a diacritic is the stacking of the two vectors to produce a final encoding vector of
length $n=37+4 = 41$. Clearly, a letter with no diacritic will have 4 zeros in the diacritic portion
of the encoding vector.

Figure~\ref{fig:One-Binary-Encoding} illustrates the three encoding schemes. The \textit{one-hot}
and \textit{binary} encoding of the whole 5-letter word \<مَرْحَبا> are illustrated as
$181 \times 5$ and $8 \times 5$ matrices respectively (Figures~\ref{fig:One-Binary-Encoding}-a,
\ref{fig:One-Binary-Encoding}-b). In Figure~\ref{fig:One-Binary-Encoding}-c only the second letter
of the word, \<بَ>, is taken an example to illustrate the \textit{two-hot} encoding. It is
obvious that the \textit{one-hot} is the most lengthy encoding; however, it is straightforward for
networks to decode since no two vectors share the same position of `1'. On the other extreme, the
\textit{binary} encoding is most economic one; however, networks may need careful design to decode
the pattern since vectors share many positions of `1's and `0's. Efficiently, the new designed
\textit{two-hot} encoding is almost 28\% of the size of \textit{one-hot} encoding.


\section{Experiments}\label{sec:model}
In this section, we explain the design and parameters of all experiments conducted in this
research. The number of experiments is the cross product of data representation parameters and
network configuration parameters.

\subsection{Data Representation Parameters}\label{sec:param-data-repr-1}
For Arabic dataset representation, there are three parameters: \textit{diacritics} (2 values),
\textit{trimming} (2 values), and \textit{encoding} (3 values); and hence there are 12 different
data representations (the $x$-axis of Figure~\ref{fig:ArabicModelsResults}). A poem can be fed to the
network with/without \textit{diacritics} (1D/0D for short); this is to study their effect on
network learning. It is anticipated that it will be much easier for the network to learn with
\textit{diacritics} since it provides more information on pronunciation and phonetics. Arabic poem
data, as indicated in Figure~\ref{fig:footn-footn-class}-a, is not balanced. To study the effect of
this unbalance, the dataset is used once with \textit{trimming} the smallest 5 meters from the
dataset and once in full (no trimming), i.e., with all 16 meters presented (1T and 0D for
short). There are three different \textit{encoding} methods, \textit{one-hot}, \textit{binary}, and
\textit{two-hot} (OneE, BinE, TwoE for short), as explained in
Sec.~\ref{sec:data-encoding}. Although all carry the same information, it is expected that a
particular encoding may be suitable for the complexity of a particular network configuration. (see
Sec.~\ref{sec:discussion} for elaboration).

For English dataset representation, there is no \textit{diacritics} and the dataset does not suffer
a severe imbalance (Figure~\ref{fig:footn-footn-class}-a). Therefore, there are just 2 different
data representations, corresponding solely to \textit{one-hot} and \textit{binary} encodings (the
$x$-axis of Figure~\ref{english_results}-a).

\subsection{Network Configuration Parameters}\label{sec:param-netw-conf}
The main Recurrent Neural Network (RNN) architectures experimented in this research are: the Long
Short Term Memory (LSTM) introduced in~\cite{Hochreiter1997LongShortTermMemory}, the Gated Recurrent
Unit (GRU)~\citep{Cho2014LearningPhraseRepresentationsUsing}, and their bidirectional variants
Bi-LSTM and Bi-GRU\@. Conceptually, GRU is almost the same as the LSTM; however, GRU has less
architectural complexity, and hence a fewer number of training parameters. From benchmarks and
literature results, it is not clear which of the four architectures is the overall winner. However,
for their comparative complexity, it can be anticipated that both LSTM and Bi-LSTM (will be always
written as (Bi-)LSTM for short) may be more accurate than their two counterparts (Bi-)GRU on much
larger datasets and vice-versa.

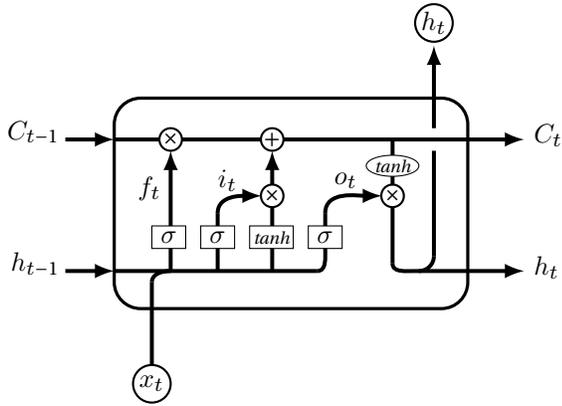
\begin{figure}[!tb]
  \centering
  \input{LSTM_unit.tex}
  \caption{Architecture of a single LSTM cell, the building block of LSTM RNN\@. (Figure adapted
    from~\cite{Colah2015UnderstandingLstmNetworks})}\label{lstm}
\end{figure}

We will give a very brief account for LSTMs, which was designed to solve the long-term dependency
problem. The other three architectures have the same design flavor and the interested reader can
refer to their literature. In theory, RNNs are capable of handling long-term dependencies. However,
in practice they do not, due to the \textit{exploding gradient} problem, where weights are updated
by the gradient of the loss function with respect to the current weights in each training epoch. In
some cases, the gradient may become infinitesimally small, which prevents weights from changing and
may stop the network from further learning. LSTMs are designed to be a remedy for this
problem. Figure~\ref{lstm} (adapted from~\cite{Colah2015UnderstandingLstmNetworks}) shows an LSTM
cell, where: $f_t$ is the forgetting gate, $i_t$ is the input gate, $o_t$ is the output gate, $C_t$
is the memory across cells, $W_j,\ U_j,\ b_j,\ j \in \left\{f, i, o\right\}$ are the weight matrices
and bias vector. The cell hidden representation $h_t$ of $x_t$ is computed as follows:%
\begin{align*}
  f_t  &= \sigma(W_f  x_t + U_f h_{t-1} + b_t),\\
  i_t  &= \sigma(W_i  x_t + U_i h_{t-1} + b_i),\\
  o_t  &= \sigma(W_o  x_t + U_o h_{t-1} + b_o),\\
  C_t  &= f_t \circ c_{t-1} + i_t \circ \tanh(W_c x_t + U_c h_{t-1} + b_c),\\
  h_t  &= o_t \circ \tanh(c_t).
\end{align*}

Next, we detail the network configuration parameters of all experiments. For Arabic dataset, there
are four parameters: \textit{cell} (2 values), \textit{layers} (2 values), \textit{size} (2
values), and \textit{weighting} (2 values). Therefore, there are $16$ different network
configurations to run on each of the 12 data representations above. This results in
$16 \times 12 (=192)$ different experiments (or models). For \textit{cell}, we tried both LSTM and
Bi-LSTM\@. Ideally, GRU and Bi-GRU should be experimented as well. However, this would require
almost the double of execution time, which would not be practical for the research life time. This
is deferred to another large scale comprehensive research currently running
(Sec.~\ref{sec:discussion}). We tried 4 and 7 \textit{layers}, with internal vectorized
\textit{size} of 50 and 82. Finally, another alternative to \textit{trimming} small classes (meters)
that was discussed above, in data representation parameters (Sec.~\ref{sec:param-data-repr-1}), is
to keep all classes but with \textit{weighting} the loss function to account for the relative class
size. For that purpose, we introduce the following \textit{weighting} function:%
\begin{align}
  w_c &= \frac{1/n_c}{\sum_{c'} 1/n_{c'}}, \label{eq:1}
\end{align}%
where $n_c$ is the sample size of class $c$, $c = 1, 2, \ldots C$, and $C$ is the total number of
classes (16 meters in our case).

For English dataset, there are four parameters: \textit{cell} (4 values), \textit{layers} (6
values), \textit{size} (4 values). We did not include \textit{weighting} since the dataset does not
suffer sever unbalance as is the case for the Arabic dataset. Therefore, there are $96$ different
network configurations to run on each of the 2 data representations above. This results in the same
number of 192 different experiments ($96 \times 2$) as those of the Arabic dataset. For
\textit{cell}, we had the luxury to experiment with the four types (Bi-)LSTM and (Bi-)GRU, since the
dataset is much smaller than the Arabic dataset. For \textit{layers}, we tried $3,4,\ldots, 8$, each
with internal vectorized \textit{size} of 30, 40, 50, and 60.

\bigskip

For all the 192 experiments on Arabic dataset and the 192 experiments on English dataset, networks
are trained using dropout of 0.2, batch size of 2048, with Adam optimizer, and 10\% for each of
validation and testing sets. Experiments are conducted on a Dell Precision T7600 Workstation with
Intel Xeon E5-2650 32x 2.8GHz CPU, 64GB RAM, 2 $\times$ NVIDIA GeForce GTX TITAN X (Pascal) GPUs;
and with: Manjaro 17.1.12 Hakoila OS, x86\_64 Linux 4.18.9-1-Manjaro Kernel.

\section{Results}\label{sec:results}
The results of all the 192 experiments on Arabic dataset and the 192 experiments on the English
dataset are presented and discussed; for each dataset, we start with the overall accuracy
followed by the individual accuracy on each class (meter).

\subsection{Results of Arabic dataset}\label{sec:arabic-results}
\subsubsection{Overall Accuracy}\label{sec:encoding-effect}
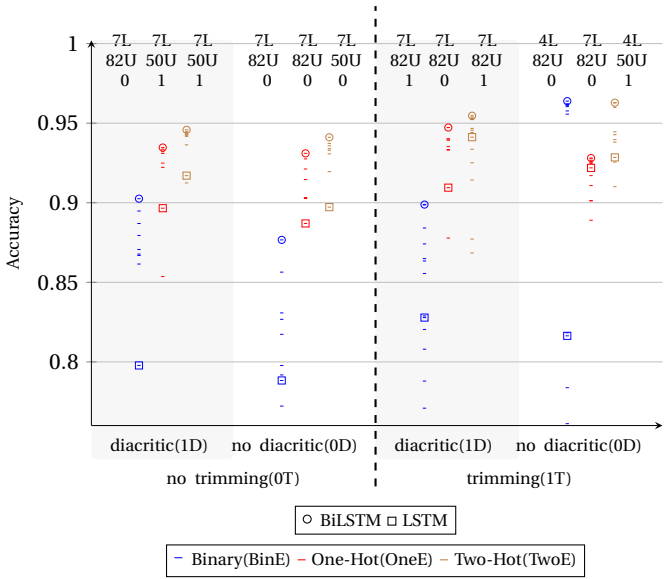
\begin{figure}[!tb]
  \centering
  \input{results_ar_models_acc.tex}
  \caption{Overall accuracy of the 192 experiments plotted as 12 vertical rug plots (one at each
    data representation: \{0T, 1T\} $\times$ \{0D, 1D\} $\times$ \{OneE, BinE, TwoE\}); each
    represents 16 exp. (for network configurations: \{4L, 7L\} $\times$ \{82U, 50U\} $\times$ \{0W,
    1W\} $\times$ \{LSTM, BiLSTM\}). For each rug plot the best model of each of the two cell
    types---(Bi-)LSTM---is labeled as circle and square respectively. BiLSTM always wins over the
    LSTM; and its network configuration parameters are listed at the top of each rug
    plot.}~\label{fig:ArabicModelsResults}
\end{figure}
First, we explain how Figure~\ref{fig:ArabicModelsResults} presents the overall accuracy of the
16 network configurations ($y$-axis) for each of the 12 data representations ($x$-axis). The
$x$-axis is divided into 4 strips corresponding to the 4 combinations of \textit{trimming} $\times$
\textit{diacritic} represented as \{0T(left), 1T(right)\} $\times$
\{0D(unshaded),1D(shaded)\}. Then, each strip includes the 3 different values of \textit{encoding}
\{BinE, OneE, TwoE\}. For each of the 12 data representations, the
$y$-axis represents a rug plot of the
accuracy of the 16 experiments; (some values are too small, and hence omitted from the
figure)\@. For each rug plot, the highest (Bi-)LSTM accuracies are labeled differently as circle and
square respectively; and the network configuration of both of them is listed at the top of the rug
plot. To explain the figure, we take as an example the most-left vertical rug plot, which
corresponds to (0T, 1D, BinE) data representation. The accuracies of the best (Bi-)LSTM are 0.9025 and
0.7978 respectively. The configuration of the former is (7L, 82U, 0W). Among all the 192
experiments, the highest
accuracy is 0.9638 and is possessed by (4L, 82U, 0W) network configuration on (1T, 0D, BinE) data
representation.

Next, we discuss the effect of each data representation and network configuration parameter on
accuracy. The effect of \textit{trimming} is clear; for particular \textit{diacritic} and
\textit{encoding}, the accuracies at 1T are consistently higher than those at 0T. E.g., the highest
accuracy at (1T, 0D, TwoE) and (0T, 0D, TwoE) are 0.9629 and 0.9411 respectively. The only
exception, with a very little difference, is (1T, 1D, BinE) vs.\ (0T, 1D, BinE). The effect of
\textit{diacritic} is obvious only at 0T (the left half of the Figure), where, at particular
\textit{encoding}, the accuracy is higher at 1D than at 0D. However, for 1T, this observation is
only true for OneE. This result is counter-intuitive if compared to what is anticipated from the
effect of diacritics. We think that this result is an artifact for the small number of network
configurations. (More on that in Sec.~\ref{sec:discussion}). The effect of \textit{encoding} is
clear as well; by looking at each individual strip out of the four strips on the $x$-axis, accuracy
is consistently highest for OneE and TwoE than BinE---the only exception is at (1T, 0D, BinE) that
performs better than the other two encodings. It seems that TwoE makes it easier for networks to
capture the patterns in data. However, we believe that there is a particular network architecture
for each encoding that is capable of capturing the same pattern with yielding the same accuracy; yet,
the set of experiments should be conducted at higher resolution of the network configuration
parameters space (Sec.~\ref{sec:discussion}).

Next, we comment on the effect of network configuration parameters. For \textit{cell} type, it is
obvious that for each data representation, the highest BiLSTM accuracy (circle) is consistently higher
than the highest LSTM accuracy (square). This is what is expected from the more complex architecture of
the BiLSTM on such a large dataset. For \textit{layers}, the more complex networks of 7 layers
achieved the highest accuracies, except for (1T, 0D, BinE) and (1T, 0D, TwoE). The straightforward
interpretation for that is the reduction in dataset size occurred by (1T, 0D) combination, which
needed less complex network. For cell \textit{size} and loss \textit{weighting}, the figure shows no
consistent effect on accuracy.

\bigskip

\subsubsection{Per-Class (Meter) Accuracy}\label{sec:per-class-accuracy}
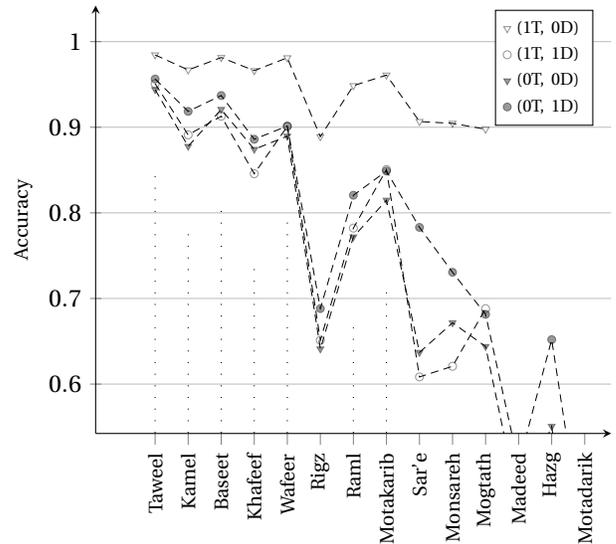
\begin{figure}[!tb]
  \centering
  \input{results_ar_per_class_accuracy.tex}
  \caption{The per-class accuracy for the best four models: \{\text{0T},\ \text{1T}\} $\times$
    \{\text{0D},\ \text{1D}\}; the $x$-axis is sorted by class size as in
    Figure~\ref{fig:footn-footn-class}. There is a descending trend with the class size, with the
    exception at \textit{Rigz} meter.}\label{fig:footn-both-models}
\end{figure}
Next, we investigate the per-class accuracy. For each of the four combinations of
\textit{trimming} $\times$ \textit{diacritic}, we select the best accuracy out of the three possible
encodings. From Figure~\ref{fig:ArabicModelsResults}, it is clear that all of them will be at TwoE,
except (1T, 0D, BinE), which is the best overall model as discussed above.

Figure~\ref{fig:footn-both-models} displays the per-class accuracy of these four models. The class
names (meters) are ordered on the $x$-axis according to their individual class size (the same order
of Figure~\ref{fig:footn-footn-class}). Several comments are in order. The overall accuracy of each of
the four models is around 0.95 (Figure~\ref{fig:ArabicModelsResults}); however, for the four models
the per-class accuracy of only 6 classes is around this value. For some classes the accuracy drops
significantly. Moreover, the common trend for the four models is that the per-class accuracy
decreases with the class size for the first 11 classes. Then, the accuracy of the two models with
\textit{trimming} keeps decreasing significantly for the remaining 5 classes. Although this
trend is associated with class size, this could only be correlations without causation. This
phenomenon, along with what was concluded above for the inconsistent effect of loss
\textit{weighting}, emphasize the importance of a more prudent design of the \textit{weighting}
function. In addition, the same full set of experiments can be re-conducted with enforcing all
classes to have equal size to assert/negate the causality assumption (Sec.~\ref{sec:discussion}).

\bigskip

\subsubsection{Encoding Effect on Learning rate and Memory
  Utilization}\label{sec:learning-rate-memory}
\begin{figure}[!tb]
  \centering
  \begin{tikzpicture}[scale=0.85]
    \input{results_ar_convergence.tex}
  \end{tikzpicture}
  \caption{Encoding effect on learning rate of the best model configurations (1T, 0D, 4L, 82U, 0W)
    with each of the three encodings.}~\label{fig:ConvergenceMemory}
\end{figure}
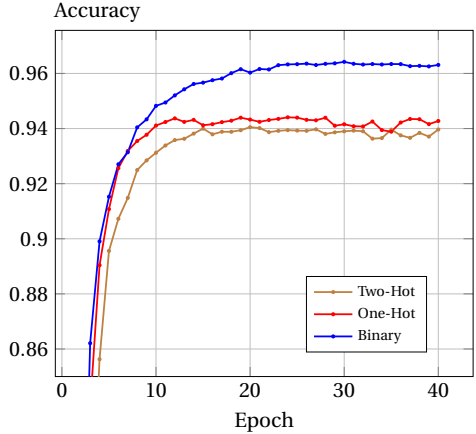
Figure~\ref{fig:ConvergenceMemory}-a shows the learning curve of the best model (4L, 82U, 0W, 1T,
0D, BinE), which converges to 0.9638, the same value displayed on
Figure~\ref{fig:ArabicModelsResults}. The Figure displays, as well, the learning curve of the same
model and parameters but with using the other two encodings. The Figure shows no big difference in
convergence speed among different encodings.

\subsection{Results of English Dataset}\label{sec:english-results}
The result presentation and interpretation for the experiments on English dataset are much easier
because of the absence of \textit{diacritic}, \textit{trimming}, and loss \textit{weighting}
parameters. The relative size of the two datasets has to be brought to attention; from
Figure~\ref{fig:footn-footn-class}, there is almost a factor of 100 in favor of the Arabic dataset.

\bigskip

\subsubsection{Overall Accuracy}\label{sec:overall-accuracy}
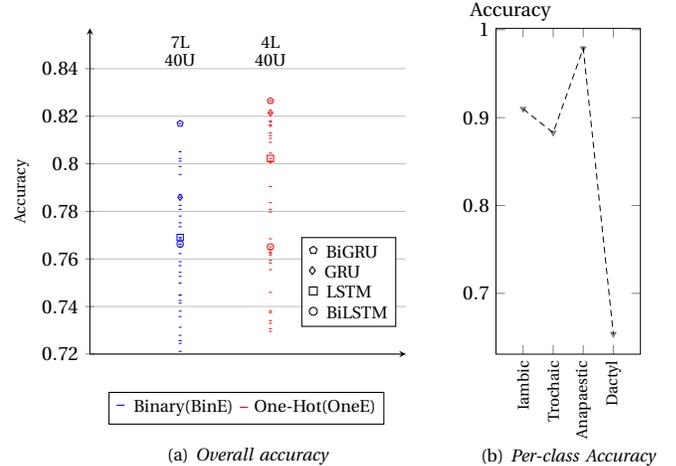
\begin{figure}[!tb]
  \centering
  \begin{tikzpicture}[scale=0.8]
    \input{results_en_models_acc.tex}
    \input{results_en_per_class_acc.tex}
  \end{tikzpicture}
  \caption{Accuracy of experiments on English dataset. (a) Overall accuracy of the 192
    experiments plotted as 2 vertical rug plots (one at each data representation: \{OneE, BinE\});
    each represents 96 exp. (for network configurations: \{3L, 4L, 5L, 6L, 7L, 8L\} $\times$ \{30U,
    40U, 50U, 60U\} $\times$ \{LSTM,\ BiLSTM,\ GRU,\ BiGRU\}). For each rug plot the best model of each
    of the four cell types---(Bi-)LSTM and (Bi-)GRU---is labeled differently. Consistently, the BiGRU
    was the winner, and its network configuration parameters are listed at the top of each rug
    plot. (b) The per-class accuracy for the best model of the 192 experiments; the $x$-axis is
    sort by the class size as in Figure~\ref{fig:footn-footn-class}. No particular trend with the
    class size is observed.}~\label{english_results}
\end{figure}

Similar to Figure~\ref{fig:ArabicModelsResults}, Figure~\ref{english_results}-a displays the accuracy of 96 network configurations ($y$-axis) for each of the 2 dataset representations
($x$-axis). The Figure shows that the highest accuracy, 0.8265, is obtained using (4L, 40U,
OneE), and BiGRU network. The \textit{encoding} is the only parameter for data representation. OneE
achieves higher accuracy than, but close to, BinE. Once again, we anticipate that experimenting
with more network configuration should resolve this difference (Sec.~\ref{sec:discussion}).

For Network configuration parameters, we start with the \textit{cell} type. At each encoding, the
best accuracy of each \textit{cell} type in descending order is: BiGRU, GRU, LSTM, then
BiLSTM\@. (Bi-)GRU models may by more suitable for this smaller size dataset. For \textit{layers},
the best models on OneE was 3L and on BinE was 7L. In contrast to the Arabic dataset, the simple 4L
achieved a better accuracy than the complex 7L, with no clear effect of cell \textit{size}. (More
discussion on that in Sec.~\ref{sec:discussion}).

\bigskip

\subsubsection{Per-Class (Meter) Accuracy}\label{sec:per-class-meter}
Figure~\ref{english_results}-b is a per-class accuracy for the best model (4L, 40U, OneE, BiGRU);
the meters are ordered on the $x$-axis descendingly with the class size as in
Figure~\ref{fig:footn-footn-class}-b. It is clear that class size is not correlated with
accuracy. Even for the smallest class, Dactyl, its size is almost one third the Iambic class
(Figure~\ref{fig:footn-footn-class}-b), which is not a huge skewing factor. A more reasonable
interpretation is this. Dactyl meter is pentameter or more; while other meters have less
repetitions. This makes Dactyl verses very distant in character space from others. And since the
network will train to optimize the overall accuracy, this may be on the expense on the class that is
both small in size and setting distant in feature space from others. (More discussion on that in
Sec.~\ref{sec:discussion}).

\section{Discussion}\label{sec:discussion}
In this section, we will elaborate on the interpretation of some results, reflect on some concepts,
and connect to the current and future research.

Sec.~\ref{sec:data-encoding} explained the three different encoding methods leveraged in this
research and cited some literature on the effect of encoding on network accuracy. Mathematically
speaking, encoding is seen as feature transformation $\mathcal{T}$, where a character $X$ is
transformed to $\mathcal{T}(X)$ in the new encoding space. Since the lossless encoding is
invertible, it is clear for any two functions (networks) and any two encodings (transformations)
that
$\eta_1\left(\mathcal{T}_1(X)\right) = \left(\eta_1\cdot\mathcal{T}_1\cdot
  \mathcal{T}_2^{-1}\right)\left(\mathcal{T}_2(X)\right) =
\eta_2\left(\mathcal{T}_2(X)\right)$. This means that if the network $\eta_1$ is the most accurate
network for the encoding $\mathcal{T}_1$, using another encoding $\mathcal{T}_2$ for the same
problem requires designing another network
$\eta_2 = \eta_1\cdot\mathcal{T}_1\cdot \mathcal{T}_2^{-1}$. However, this network may be of
complicated architecture to be able to ``decode'' a terse or complex pattern $\mathcal{T}_2(X)$. The
behavior of the three encodings BinE, OneE, and TwoE in this paper can be seen in the light of this
discussion. The most terse representation is the BinE ($n=8$) and the most sparse representation is
the OneE ($n=181$); and in between comes our TwoE ($n=41$) as a smart design and compromise between
the low dimensionality of BinE and the self-decoded nature of the OneE
(Sec.~\ref{sec:data-encoding}). This may be a qualitative interpretation to why the accuracy of the
best models was always possessed by the TwoE, yet with one exception at the BinE
(Sec.~\ref{sec:arabic-results}). However, from Figures~\ref{fig:ArabicModelsResults} and
\ref{english_results}, the rug plots reveal that the populations of accuracy at different
encodings do interleave and each encoding can perform better than others at some experiments. We
emphasize that this effect is an artifact to the non exhaustive network configuration parameters and
experiments conducted in this research. Had we covered the configuration parameter space then all
encoding methods would produce the same accuracy, yet at different network architectures, as each
encoding requires the right network architecture to learn from (or to ``decode'').

Sec.~\ref{sec:param-netw-conf} detailed the network configuration parameters for both Arabic
datasets (\{4L, 7L\} $\times$ \{82U, 50U\} $\times$ \{0W, 1W\} $\times$ \{LSTM, BiLSTM\} = 16
networks) and for English dataset (\{3L, 4L, 5L, 6L, 7L, 8L\} $\times$ \{30U, 40U, 50U, 6U\}
$\times$ \{LSTM,\ BiLSTM,\ GRU,\ BiGRU\} = 96 networks). Each experiment runs almost in one hour (30
epochs $\times$ 2 min/epoch) on the mentioned hardware (Sec.~\ref{sec:model}). The total run time of
all network configurations on all data representations for both Arabic and English datasets was
$16\times 12+96\times 2 = 384$ hours, i.e., more than two weeks! We are currently working on more
exhaustive set of experiments to cover a good span of the network configuration parameter space to
both confirm the above discussion on encoding and to boost the per-class accuracy on both datasets.

\bigskip

The per-class accuracy for both datasets needs investigation; in particular, the interesting trend
between the per-class accuracy and the class size of the Arabic dataset needs more investigation. We
speculate that this is a mere correlation that does not imply causation; and the reason for this
trend may be attributed to the difficulty of, or the similarity between, the meters having small
class size. This difficulty, or similarity, may be what is responsible for the low accuracy
(Figure~\ref{fig:footn-both-models}) on a hand, and the lack of interest of poets to compose at
these meters, which resulted in their scarcity (Figure~\ref{fig:footn-footn-class}), on the other
hand.

Diacritic effect is explained in Sec.~\ref{sec:results}; experiments with diacritics scored higher
than those without diacritics only when small class size were trimmed from the datasets (1T). When
including the whole dataset (0T) the effect of diacritics was not consistent. This interesting
phenomenon needs more investigation, since the phonetic pattern of any meter is uniquely identified
by diacritics (Sec.~\ref{sec:arab-poetry-text}). This may be connected to the observation above of
the per-class accuracy.

For more investigation of both phenomena, we are working on a randomized-test-like experiments in
which all classes will be forced to have equal size $n$. We will study how the per-class accuracy or
overall accuracy, along with their two individual components (precision and recall), behave and how
the diacritic effect changes in terms of both $n$ and the number of involved classes $k$, where
$2 \leq k \leq K$, and $K(=16)$ is the total number of meters.

\section{Conclusion}\label{sec:conclusion}
This paper aimed at training Recurrent Neural Networks (RNN) at the character level on Arabic and
English written poem to learn and recognize their meters that make poem sounding rhetoric or
phonetic when pronounced. This can be considered a step forward for language understanding,
synthesis, and style recognition. The datasets were crawled from several non technical online
sources; then cleaned, structured, and published to a repository that is made publicly available for
scientific research. To the best of our knowledge, using Machine Learning (ML) in general and Deep
Neural Networks (DNN) in particular for learning poem meters and phonetic style from written text,
along with the availability of such a dataset, is new to literature.

For the computational intensive nature and time complexity of RNN training, our network
configurations were not exhaustive to cover a very wide span of training parameter configurations
(e.g., number of layers, cell size, etc). Nevertheless, the classification accuracy obtained on the
Arabic dataset was remarkable, specially if compared to that obtained from the deterministic and
human derived rule-based algorithms available in literature. However, the door is opened to many
questions and more exploration; to list a few: how to increase the accuracy on English dataset, why diacritic
effect is not consistent, and why some meters possess low per-class accuracy.

\section{Acknowledgment}\label{sec:acknowledgment}
The authors gratefully acknowledge the support of NVIDIA Corporation with the donation of the two
Titan X Pascal GPU used for this research. The authors are grateful, as well, to each of Ali
H. El-Kassas, Ali O. Hassan, Abdallah R. Albohy, and Ahmed A. Abouelkahire for their contribution to
an early version of this work.

\footnotesize
\bibliographystyle{IEEEtran}
\bibliography{TempBib.bib}

\end{document}

%% file: pgfplot_configurations.tex
\pgfplotsset{%
  pointModelsFiguresStyle/.style args={#1}{%
    color=#1,
    mark=0,
    only marks,
    mark size=1.5pt,
    point meta=explicit symbolic,
  }
}

\pgfplotsset{%
  pointRug/.style args={#1}{%
    color=#1,
    mark=-,
    only marks,
    mark size=1.5pt,
    point meta=explicit symbolic,
  }
}

\pgfplotsset{%
  pointBiLSTM/.style args={#1}{%
    color=#1,
    mark=o,
    only marks,
    mark size=1.5pt,
    point meta=explicit symbolic,
  }
}

\pgfplotsset{%
  pointLSTM/.style args={#1}{%
    color=#1,
    mark=square,
    only marks,
    mark size=1.5pt,
    point meta=explicit symbolic,
  }
}

\pgfplotsset{%
  pointBiGRU/.style args={#1}{%
    color=#1,
    mark=pentagon,
    only marks,
    mark size=1.5pt,
    point meta=explicit symbolic,
  }
}

\pgfplotsset{%
  pointGRU/.style args={#1}{%
    color=#1,
    mark=diamond,
    only marks,
    mark size=1.5pt,
    point meta=explicit symbolic,
  }
}

%% file: dataset_size_ar.tex
\begin{axis}[
  symbolic x coords={Taweel,
    Kamel,
    Baseet,
    Khafeef,
    Wafeer,
    Rigz,
    Raml,
    Motakarib,
    Sar'e,
    Monsareh,
    Mogtath,
    Madeed,
    Hazg,
    Motadarik,
    Moktadib,
    Modar'e
  },
  xtick=data,
  every axis y label/.style= {at={( 0.15, 1.07)}, anchor=north},
  ylabel style={font=\footnotesize},
  xticklabel style = {font=\footnotesize},
  ylabel={Verses},
  height=8cm,
  x=0.4cm,
  x tick label style={rotate=90, anchor=east},
  enlarge y limits=0.07,
  name=left plot,%
  title=(a) \textit{Arabic Dataset},%
  title style={at={(0.5,-.4)}}%
  ]
  \addplot[ybar,color=black,mark=*, only marks,
  point meta=explicit symbolic] coordinates {
    (Taweel, 416428)
    (Kamel,  370116)
    (Baseet, 244583)
    (Khafeef,     157880)
    (Wafeer,     143148)
    (Rigz,     119286)
    (Raml,     79560)
    (Motakarib,     63613)
    (Sar'e,     59370)
    (Monsareh,     28768)
    (Mogtath,     18062)
    (Madeed,     7808)
    (Hazg,     7468)
    (Motadarik,     5144)
    (Moktadib,     799 )
    (Modar'e,     288 )
  };

  \draw[loosely dotted] (axis cs:Taweel, 0) -- (axis cs:Taweel, 416428);
  \draw[loosely dotted] (axis cs:Kamel,  0) -- (axis cs:Kamel,  370116);
  \draw[loosely dotted] (axis cs:Baseet, 0) -- (axis cs:Baseet, 244583);
  \draw[loosely dotted] (axis cs:Khafeef,0) -- (axis cs:Khafeef,157880);
  \draw[loosely dotted] (axis cs:Wafeer, 0) -- (axis cs:Wafeer, 143148);
  \draw[loosely dotted] (axis cs:Rigz,   0) -- (axis cs:Rigz,   119286);
  \draw[loosely dotted] (axis cs:Raml,   0) -- (axis cs:Raml,   79560);
  \draw[loosely dotted] (axis cs:Motakarib, 0) -- (axis cs:Motakarib, 63613);
  \draw[loosely dotted] (axis cs:Sar'e, 0)   -- (axis cs:Sar'e, 59370);
  \draw[loosely dotted] (axis cs:Monsareh, 0) -- (axis cs:Monsareh,28768);
  \draw[loosely dotted] (axis cs:Mogtath, 0) -- (axis cs:Mogtath, 18062);
  \draw[loosely dotted] (axis cs:Madeed,  0) -- (axis cs:Madeed,  7808);
  \draw[loosely dotted] (axis cs:Hazg,    0) -- (axis cs:Hazg,    7468);
  \draw[loosely dotted] (axis cs:Motadarik,0) -- (axis cs:Motadarik, 5144);
  \draw[loosely dotted] (axis cs:Moktadib, 0) -- (axis cs:Moktadib,  799 );
  \draw[loosely dotted] (axis cs:Modar'e,  0) -- (axis cs:Modar'e,   288 );

\end{axis}


%% file: dataset_size_en.tex
\begin{axis}[
    ylabel={Verses},
        symbolic x coords={Iambic, Trochee, Anapaestic, Dactyl},
        xtick=data,
        every axis y label/.style= {at={( 0.5, 1.07)}, anchor=north},
        ylabel style={font=\footnotesize},
        xticklabel style = {font=\footnotesize},
        height=8cm,
        x=0.5cm,
        enlarge x limits=0.3,
        enlarge y limits=0.07,
        nodes near coords,
        x tick label style={rotate=90, anchor=east},
        y tick label style={scaled ticks=base 10:-3},
        ymin=0,
        at={($(left plot.south east)+(0.7cm,0)$)},%
        title=(b) \textit{English Dataset},%
        title style={at={(0.5,-.4)}}%
      ]

  \addplot[mark=*, only marks,
    point meta=explicit symbolic] coordinates {
    (Trochee,    5418)
    (Anapaestic, 5378)
    (Dactyl,     1397)
    (Iambic,     5550)
};

    \draw[loosely dotted] (axis cs:Trochee, 0) -- (axis cs:Trochee, 5418);
    \draw[loosely dotted] (axis cs:Anapaestic, 0) -- (axis cs:Anapaestic, 5378);
    \draw[loosely dotted] (axis cs:Dactyl, 0) -- (axis cs:Dactyl, 1397);
    \draw[loosely dotted] (axis cs:Iambic, 0) -- (axis cs:Iambic, 5550);
\end{axis}%


%% file: encoding_three_figures_together.tex
\begin{tikzpicture}
    \node at (-12,0){\input{encoding_onehot.tex}};
    \node at (-7,0){\input{encoding_binary}};
    \node at (0,0){\input{encoding_twohot.tex}};
    \node at (-12+0.3,-3){\small (a) \textit{One-Hot}};
    \node at (-7+0.3, -3){\small (b) \textit{Binary}};
    \node at (0-0.5,  -3){\small (c) \textit{Two-Hot}};
\end{tikzpicture}


%% file: encoding_onehot.tex
\begin{tikzpicture}   
  \def \x {0.57} 
  \def \s {1} 
  \newcommand\letters{{"ا","بَ","حَ", "رْ", "مَ"}}

  \def \length {2.95}
  \draw[rounded corners=1pt] (\s+\x*0, 0) rectangle (\s+0.5+\x*0, \length) node[above, xshift=-0.25cm] {\<مَ>};
  \draw[rounded corners=1pt] (\s+\x*1, 0) rectangle (\s+0.5+\x*1, \length) node[above, xshift=-0.25cm] {\<رْ>};
  \draw[rounded corners=1pt] (\s+\x*2, 0) rectangle (\s+0.5+\x*2, \length) node[above, xshift=-0.25cm] {\<حَ>};
  \draw[rounded corners=1pt] (\s+\x*3, 0) rectangle (\s+0.5+\x*3, \length) node[above, xshift=-0.25cm] {\<بَ>};
  \draw[rounded corners=1pt] (\s+\x*4, 0) rectangle (\s+0.5+\x*4, \length) node[above, xshift=-0.25cm] {\<ا>};

  \def \d {3.9 - 0.15}
  \draw (\s+0.25-0.007, 3.9 ) -- (\s+2.53+0.007, 3.9 ); 
  \draw (\s+0.25, 3.9 ) -- (\s+0.25, \d ); 
  \draw (\s+0.25*3+0.03, 3.9 ) -- (\s+0.25*3+0.03, \d ); 
  \draw (\s+0.25*5+0.15, 3.9+0.15 ) -- (\s+0.25*5+0.15, \d ) 
  node[above, yshift=0.3cm]{\<مَرْحَبَا>}; 
  \draw (\s+0.25*7+0.2, 3.9 ) -- (\s+0.25*7+0.2, \d ); 
  \draw (\s+2.53, 3.9 )  -- (\s+2.53, \d );  

  \node at (\s+0.25, \length - 0.25)    {1}; 
  \node at (\s+0.25, \length - 0.6)     {0};
  \node at (\s+0.25, \length - 0.95)    {0};
  \node at (\s+0.25, \length - 1.3)     {0};
  \node at (\s+0.25,  1.1)     {\vdots};
  \node at (\s+0.25, \length - 2.7)     {0};

  \def \d {0.07}
  \node at (\s+0.25*3+\d, \length - 0.25)    {0}; 
  \node at (\s+0.25*3+\d, \length - 0.6)     {\vdots};
  \node at (\s+0.25*3+\d, \length - 1.18)     {1};
  \node at (\s+0.25*3+\d, 1.28) {\vdots};
  \node at (\s+0.25*3+\d, \length - 2.35)    {0};
  \node at (\s+0.25*3+\d, 0.25) {0}; 

  \node at (\s+0.25*5+\d*2, \length - 0.25)    {0}; 
  \node at (\s+0.25*5+\d*2, \length - 0.6)     {\vdots};
  \node at (\s+0.25*5+\d*2, \length - 1.35)     {1};
  \node at (\s+0.25*5+\d*2, 1.18) {\vdots};
  \node at (\s+0.25*5+\d*2, \length - 2.35)    {0};
  \node at (\s+0.25*5+\d*2, 0.25) {0}; 

  \node at (\s+0.25*7+\d*3, \length - 0.25)    {0}; 
  \node at (\s+0.25*7+\d*3, \length - 0.6)     {0};
  \node at (\s+0.25*7+\d*3, \length - 0.95)     {0};
  \node at (\s+0.25*7+\d*3, \length - 1.33)     {\vdots};
  \node at (\s+0.25*7+\d*3, 1.07) {1};
  \node at (\s+0.25*7+\d*3, \length - 2.2)    {\vdots};
  \node at (\s+0.25*7+\d*3, 0.25) {0}; 

  \node at (\s+0.25*9+\d*4, 3 - 0.25) {0}; 
  \node at (\s+0.25*9+\d*4, 3 - 0.69) {0};
  \node at (\s+0.25*9+\d*4, 0.25) {0};
  \node at (\s+0.25*9+\d*4, 0.25*4 - 0.1) {\vdots};
  \node at (\s+0.25*9+\d*4, 0.9 + 0.4) {1};
  \node at (\s+0.25*9+\d*4, 1.3 + 0.6) {\vdots};

  \def \d {0.6} 
  \draw[arrows=->] (\d, 1.7) -- (\d, 2.9); 
  \draw[arrows=<-] (\d, 0.1) -- (\d, 1.1) node[above, yshift=0.07cm]
  {\small{181}};

\end{tikzpicture}%


%% file: encoding_binary.tex
\begin{tikzpicture}[x=1cm,y=1cm]

  \def \x {0.57}
  \def \s {1} 
  \newcommand\letters{{\<ا>,\<بَ>,\<حَ>,\<رْ>,>}}


  \def \length {2.95}
  \draw[rounded corners=1pt] (\s+\x*0, 0) rectangle (\s+0.5+\x*0, \length) node[above, xshift=-0.25cm] {\<مَ>};
  \draw[rounded corners=1pt] (\s+\x*1, 0) rectangle (\s+0.5+\x*1, \length) node[above, xshift=-0.25cm] {\<رْ>};
  \draw[rounded corners=1pt] (\s+\x*2, 0) rectangle (\s+0.5+\x*2, \length) node[above, xshift=-0.25cm] {\<حَ>};
  \draw[rounded corners=1pt] (\s+\x*3, 0) rectangle (\s+0.5+\x*3, \length) node[above, xshift=-0.25cm] {\<بَ>};
  \draw[rounded corners=1pt] (\s+\x*4, 0) rectangle (\s+0.5+\x*4, \length) node[above, xshift=-0.25cm] {\<ا>};

  \def \d {3.9 - 0.15}
  \def \z {0.2}
  \draw (\s+0.25-0.007, 3.9 ) -- (\s+2.53+0.007, 3.9 ); 
  \draw (\s+0.25, 3.9 ) -- (\s+0.25, \d ); 
  \draw (\s+0.25*3+0.03, 3.9 ) -- (\s+0.25*3+0.03, \d ); 
  \draw (\s+0.25*5+0.15, 3.9+0.15 ) -- (\s+0.25*5+0.15, \d ) 
  node[above, yshift=0.3cm]{\<مَرْحَبَا>}; 
  \draw (\s+0.25*7+0.2, 3.9 ) -- (\s+0.25*7+0.2, \d ); 
  \draw (\s+2.53, 3.9 )  -- (\s+2.53, \d );  

  \node at (\s+0.25, \length - 0.25)    {0}; 
  \node at (\s+0.25, \length - 0.6)     {0};
  \node at (\s+0.25, \length - 0.95)    {0};
  \node at (\s+0.25, \length - 1.3)     {0};
  \node at (\s+0.25, \length - 1.65)    {0};
  \node at (\s+0.25, \length - 2)       {0};
  \node at (\s+0.25, \length - 2.35)    {0};
  \node at (\s+0.25, \length - 2.7)     {1};

  \def \d {0.07}
  \node at (\s+0.25*3+\d*1, \length - 0.25)    {0}; 
  \node at (\s+0.25*3+\d*1, \length - 0.6)     {0};
  \node at (\s+0.25*3+\d*1, \length - 0.95)    {1};
  \node at (\s+0.25*3+\d*1, \length - 1.3)     {0};
  \node at (\s+0.25*3+\d*1, \length - 1.65)    {0};
  \node at (\s+0.25*3+\d*1, \length - 2)       {1};
  \node at (\s+0.25*3+\d*1, \length - 2.35)    {1};
  \node at (\s+0.25*3+\d*1, \length - 2.7)     {1};

  \node at (\s+0.25*5+\d*2, \length - 0.25)    {0}; 
  \node at (\s+0.25*5+\d*2, \length - 0.6)     {0};
  \node at (\s+0.25*5+\d*2, \length - 0.95)    {1};
  \node at (\s+0.25*5+\d*2, \length - 1.3)     {0};
  \node at (\s+0.25*5+\d*2, \length - 1.65)    {1};
  \node at (\s+0.25*5+\d*2, \length - 2)       {0};
  \node at (\s+0.25*5+\d*2, \length - 2.35)    {1};
  \node at (\s+0.25*5+\d*2, \length - 2.7)     {1};

  \node at (\s+0.25*7+\d*3, \length - 0.25)    {1}; 
  \node at (\s+0.25*7+\d*3, \length - 0.6)     {0};
  \node at (\s+0.25*7+\d*3, \length - 0.95)    {0};
  \node at (\s+0.25*7+\d*3, \length - 1.3)     {1};
  \node at (\s+0.25*7+\d*3, \length - 1.65)    {1};
  \node at (\s+0.25*7+\d*3, \length - 2)       {0};
  \node at (\s+0.25*7+\d*3, \length - 2.35)    {1};
  \node at (\s+0.25*7+\d*3, \length - 2.7)     {1};

  \node at (\s+0.25*9+\d*4, \length - 0.25)    {0}; 
  \node at (\s+0.25*9+\d*4, \length - 0.6)     {0};
  \node at (\s+0.25*9+\d*4, \length - 0.95)    {1};
  \node at (\s+0.25*9+\d*4, \length - 1.3)     {1};
  \node at (\s+0.25*9+\d*4, \length - 1.65)    {1};
  \node at (\s+0.25*9+\d*4, \length - 2)       {1};
  \node at (\s+0.25*9+\d*4, \length - 2.35)    {0};
  \node at (\s+0.25*9+\d*4, \length - 2.7)     {1};

  \def \vColumn {0.6} 
  \draw[arrows=->] (\vColumn, 1.7) -- (\vColumn, 2.9); 
  \draw[arrows=<-] (\vColumn, 0.1) -- (\vColumn, 1.1) node[above, yshift=0.07cm] {\small{8}};


\end{tikzpicture}


%% file: encoding_twohot.tex
\newsavebox{\columnVector}
\savebox{\columnVector}{$\left[\begin{smallmatrix}k\\m\end{smallmatrix}\right]$}

\begin{tikzpicture}


  \def \recWidthA {0.5}
  \def \yStartA   {1.5}
  \def \yEndA     {3}
  \def \xStartA   {1.5}
  \def \xEndA     {0.5}
  \def \xStartB   {0}
  \def \yStartB   {1}
  \def \xEndB     {2}
  \def \yEndB     {3.5}
  \def \half      {\yEndA/2 - \yStartA/2}
  \def \xStartC   {3 +0.5}
  \def \yStartC   {0.5 -.1}
  \def \xEndC     {3.5 +0.5}
  \def \yEndC     {4 +.1 }
  \def \shiftMargin {0.3}
  \newcommand\numbers{{0.25, 0.6, 0.95, 1.3, 1.65, 2, 2.35, 2.7}}

  \draw[rounded corners=2pt] (\xStartA, \yStartA) rectangle (\xStartA + \recWidthA, \yEndA)
  node [above, xshift=-.25cm] {\<◌َ>};
  \node at (0.25 + \xStartA, 3 -\numbers[0]) {1};
  \foreach \j in {1,...,3}
  \node at (0.25 + \xStartA, \yEndA -\numbers[\j]) {0};

  \node at (\xStartA-\shiftMargin, \yStartA + \half) {\small 4};
  \draw[arrows=-angle 90, very thin] (\xStartA-\shiftMargin, \yStartA + \half +.3) -- (\xStartA-\shiftMargin, \yEndA -0.1);
  \draw[arrows=angle 90-, very thin] (\xStartA-\shiftMargin, \yStartA + .1) -- (\xStartA-\shiftMargin, \yStartA + \half -.3);

  \draw[rounded corners=2pt] (\xStartB +0.5, \yStartB) rectangle (\yStartB +0.5 -1.5, \yEndB)
  node[above, xshift=0.25cm] {\<ب>};
  \node at (\xStartB +0.25, \yEndB -\numbers[0]) {0};
  \node at (\xStartB +0.25, \yEndB -\numbers[1]) {1};
  \node at (\xStartB +0.25, \yEndB -\numbers[2]) {0};
  \node at (\xStartB +0.25, \yEndB -\numbers[3]) {0};
  \node at (\xStartB +0.25, \yEndB -\numbers[4] ) {\vdots};
  \node at (\xStartB +0.25, \yStartB +\numbers[0]) {0};

  \node at (-\shiftMargin, \yStartA + \half) {\small 37};
  \draw[arrows=-angle 90, very thin] (-\shiftMargin, \yStartA + \half +.3) -- (-\shiftMargin, \yEndB -0.1);
  \draw[arrows=angle 90-, very thin] (-\shiftMargin, \yStartB + .1) -- (-\shiftMargin, \yStartA + \half -.3);

  \node at (0.8, \yStartA +\half) {$+$};
  \node at (2.5 + .02, \yStartA +\half) {$=$};

  \draw[rounded corners=2pt]  (\xStartC, \yStartC) rectangle (\xEndC, \yEndC)
  node[above, xshift=-0.25cm] {\<بَ>};
  ;
  \node at (\xStartC +0.25, \yEndC -\numbers[0]) {1};
  \node at (\xStartC +0.25, \yEndC -\numbers[1]) {0};
  \node at (\xStartC +0.25, \yEndC -\numbers[2]) {0};
  \node at (\xStartC +0.25, \yEndC -\numbers[3]) {0};

  \node at (\xStartC +0.25, \yEndC -\numbers[4] -.2) {0};
  \node at (\xStartC +0.25, \yEndC -\numbers[5] -.2) {1};
  \node at (\xStartC +0.25, \yEndC -\numbers[6] -.2) {0};
  \node at (\xStartC +0.25, \yEndC -\numbers[7] -.2) {\vdots};
  \node at (\xStartC +0.25, \yStartC +\numbers[0]) {0};

  \node at (\xStartC -\shiftMargin, \yStartA +\half) {\small 41};
  \draw[arrows=-angle 90, very thin]  (\xStartC -\shiftMargin, 2.5) -- (\xStartC -\shiftMargin, \yEndC -0.1);
  \draw[arrows=angle 90-, very thin]  (\xStartC -\shiftMargin, 0.5) -- (\xStartC -\shiftMargin, 2);

  \draw [decorate,decoration={brace,amplitude=2pt,mirror,raise=2pt}, very thin]
  (\xEndC +\shiftMargin -.1, 2.5 +.2) -- (\xEndC +\shiftMargin  -.1, \yEndC -0.1);

  \draw [decorate,decoration={brace,amplitude=2pt,mirror,raise=2pt}, very thin]
  (\xEndC +\shiftMargin -.1, 0.5) -- (\xEndC +\shiftMargin -.1, 2.4);

  \node [rectangle, text width= 5em,font=\footnotesize] at (5.6, 3.4)
    {\scriptsize $4 \times 1$ encoding vector representing diacritic \<◌َ>};

  \node [rectangle, text width= 5em,font=\footnotesize] at (5.6, 1.4)
    {\scriptsize $37 \times 1$ encoding vector representing letter \<ب>};

\end{tikzpicture}


%% file: LSTM_unit.tex
\begin{tikzpicture}


  \def \yONE {0.5}
  \def \yTWO {3.3}
  \draw[rounded corners=10pt, very thick]  (1.5, 0.5) rectangle (6.2, 3.3);

  \def \sigmoidWidth  {0.45}
  \def \sigmoidInDist {0.2}
  \def \sigmoidHeight  {0.3}
  \def \yB  {1.3}
  \def \yD  {2.75}
  \def \shift {0.3}
  \draw  (2, \yB) rectangle (2+ \sigmoidWidth, \yB + \sigmoidHeight) node[pos=0.5] {$\sigma$};

  \def \r {.15cm}
  \draw[thick] (2.25, \yD) circle (\r) node {$\times$};

  \def \xS {2.45 +\sigmoidInDist}
  \draw  (\xS, \yB) rectangle (\xS +\sigmoidWidth, \yB + \sigmoidHeight) node[pos=0.5] {$\sigma$};

  \def \xSS {\xS +\sigmoidWidth + \sigmoidInDist}
  \draw  (\xSS, \yB) 
  rectangle 
  (\xSS +\sigmoidWidth +0.13, \yB +\sigmoidHeight)
  node[pos=0.5] {\scriptsize \textit{tanh}};

  \def \xSSS {\xSS +\sigmoidWidth +0.13 +\sigmoidInDist}
  \draw  (\xSSS, \yB) rectangle (\xSSS +\sigmoidWidth, \yB +\sigmoidHeight)
  node[pos=0.5] { $\sigma$};

  \def \xTimesB {3.6}
  \def \yC      {2}

  \draw[thick] (\xTimesB, \yC) circle (\r) node {$\times$};
  \draw[thick] (\xTimesB, \yD) circle (\r) node {$+$};

  \draw[thick] (\xTimesB +1.3 +\shift, \yC) circle (\r) node {$\times$};
  \draw (\xTimesB +1.3 +\shift, \yC +.4)  
  ellipse (.35cm and .14cm) 
  node {\scriptsize \textit{tanh}};

  \draw[arrows=-latex, line width=1.5pt]  (0.85, \yD) -- (1.5, \yD) 
  node[left, xshift=-.6cm, yshift=0.1cm] {$C_{t-1}$};
  \draw[line width=1.5pt]  (1.5, \yD) -- (2.1, \yD);
  \draw[line width=1.5pt]  (2.4, \yD) -- (3.46, \yD);
  \draw[arrows=-latex, line width=1.5pt]  (3.75, \yD) -- (6.75 +0.2, \yD)
  node[right, xshift=0.0cm, yshift=0.05cm] {$C_t$};

  \draw[arrows=-latex, line width=1.5pt]  (2.25, \yB +\sigmoidHeight) -- (2.25, 2.6)
  node[left, yshift=-0.5cm] {$f_t$};

  \def \levelB {1}
  \draw[arrows=-latex, line width=1.5pt] (0.85, \levelB) -- (1.5, \levelB)
  node[left, xshift=-.6cm, yshift=0.1cm] {$h_{t-1}$};

  \draw[line width=1.5pt] (1.5, \levelB) 
  --
  (4.2, \levelB);

  \def \xOfThirdSigmoid {\xSSS + \sigmoidWidth/2}
  \draw[line width=1.5pt] (4.2, \levelB) to[out=0,in=270] (\xOfThirdSigmoid, \yB);

  \draw[line width=1.5pt] (\xOfThirdSigmoid, \yB + \sigmoidHeight) 
  --
  (\xOfThirdSigmoid, \yB + \sigmoidHeight + 0.17) ;

  \draw[arrows=-latex, line width=1.5pt] (\xOfThirdSigmoid, \yB + \sigmoidHeight+0.17) 
  to[out=90, in=180] 
  (\xTimesB +1.3 -0.14 +\shift , \yC) 
  node [left, yshift=0.2cm, xshift=-0.2cm] {$o_t$};

  \draw [line width=1.5pt] (\xTimesB, \levelB)
  --
  (\xTimesB, \yB);

  \draw [line width=1.5pt] (\xTimesB, \yB + \sigmoidHeight)
  --
  (\xTimesB, \yB + \sigmoidHeight +0.24);

  \draw [arrows=-latex, line width=1.5pt] (\xTimesB, 2.15)
  --
  (\xTimesB, 2.6);

  \draw[line width=1.5pt] (2.25, \levelB)
  --
  (2.25, \yB);
  
  \draw[line width=1.5pt] (\xS +\sigmoidWidth/2, \levelB)
  --
  (\xS +\sigmoidWidth/2, \yB);

  \draw[line width=1.5pt] (\xS +\sigmoidWidth/2, \yB +\sigmoidHeight)
  -- 
  (\xS +\sigmoidWidth/2, \yB +\sigmoidHeight + 0.17);

  \draw[arrows=-latex, line width=1.5pt] (\xS +\sigmoidWidth/2, \yB + \sigmoidHeight+0.17) 
  to[out=90, in=180] 
  (\xTimesB-0.14, \yC)
  node [left, yshift=0.2cm, xshift=-0.2cm] {$i_t$};

  \def \bigR {0.25cm}
  \def \C    {1}
  \def \xBigCicleONE {2}
  \def \xBigCicleTWO {5.5 + 0.25}
  \draw[thick] (\xBigCicleONE, \yONE - \C) circle (\bigR) node {$x_t$};
  \draw[thick] (\xBigCicleTWO, \yTWO + \C) circle (\bigR) node {$h_t$};

  \draw[line width=1.5pt] (\xBigCicleONE, \yONE -\C +0.25) 
  --
  (\xBigCicleONE, \yONE -\C +0.25 + 1.1);

  \draw[line width=1.5pt] (\xBigCicleONE, \yONE -\C +0.25 + 1.1)
  to[out=90, in=180]
  (2.25, 1);

  \draw[line width=1.5pt] (\xTimesB +1.3 +\shift, 1.1)
  --
  (\xTimesB +1.3 +\shift, \yC - 0.15);

  \draw[arrows=-latex, line width=1.5pt] (5.5, 1) 
  --
  (6.75 +0.2, 1)
  node[right, xshift=0.0cm, yshift=0.05cm] {$h_t$};

  \draw[line width=1.5pt] (\xTimesB +1.3 +\shift, 1.1)
  to[out=270, in=180]
  (5.5, 1);

  \draw[line width=1.5pt] (\xTimesB +1.3 +\shift, 2.14)
  --
  (\xTimesB +1.3 +\shift, 2.26);

  \draw[line width=1.5pt] (\xTimesB +1.3 +\shift, \yD)
  --
  (\xTimesB +1.3 +\shift, 2.54);

  \draw[line width=1.5pt] (\xBigCicleTWO, \levelB +0.2)
  --
  (\xBigCicleTWO, 2.6);

  \draw[line width=1.5pt] (\xBigCicleTWO - 0.2, \levelB)
  to[out=0, in=270]
  (\xBigCicleTWO, \levelB +0.2);

  \draw[arrows=-latex, line width=1.5pt]
  (\xBigCicleTWO, \yD +0.15)
  --
  (\xBigCicleTWO, 4);

\end{tikzpicture}


%% file: results_ar_models_acc.tex
\begin{tikzpicture}[scale=0.90]


  \def \maxHeight{5.7}
  \def \yONE{-0.8}
  \def \yTWO{-0.3}

  \fill[gray!30, opacity=0.2, rounded corners=2pt] (0,-0.5) rectangle (2.1,\maxHeight);
  \fill[gray!30, opacity=0.2, rounded corners=2pt] (4.2,-0.5) rectangle (6.32,\maxHeight);

  \draw[dashed, thick] (4.2, 0 - 0.9) -- (4.2, \maxHeight +0.5);

  \node [align=center, text width=3cm, inner sep=0.25cm] at (2.1, \yONE) {\scriptsize no trimming(0T)};
  \node [align=center, text width=4cm, inner sep=0.25cm] at (6.32,\yONE) {\scriptsize trimming(1T)};

  \node [align=center, text width=3cm, inner sep=0.25cm] at (1, \yTWO) {\scriptsize diacritic(1D)};
  \node [align=center, text width=3cm, inner sep=0.25cm] at (3, \yTWO) {\scriptsize no diacritic(0D)};
  \node [align=center, text width=3cm, inner sep=0.25cm] at (5+0.2, \yTWO) {\scriptsize diacritic(1D)};
  \node [align=center, text width=3cm, inner sep=0.25cm] at (7+0.2, \yTWO) {\scriptsize no diacritic(0D)};

  \def \layerHeight{5.7}
  \def \unitHeight{\layerHeight - 0.3}
  \def \weightedHeihgt{\layerHeight - 0.6}

  \def \step{0.2}
  \def \move{0.1}

  \node  at (3.5*   \step -0.15-\move, \layerHeight) {\scriptsize 7L};
  \node  at (3.5*   \step -0.15-\move, \unitHeight) {\scriptsize 82U};
  \node  at (3.5*   \step -0.1 -\move, \weightedHeihgt) {\scriptsize 0};

  \node  at (5.2*   \step, \layerHeight) {\scriptsize 7L};
  \node  at (5.2*   \step, \unitHeight) {\scriptsize 50U};
  \node  at (5.2*   \step, \weightedHeihgt) {\scriptsize 1};

  \node  at (7*     \step +0.15+\move, \layerHeight) {\scriptsize 7L};
  \node  at (7*   \step   +0.15+\move, \unitHeight) {\scriptsize 50U};
  \node  at (7*   \step   +0.1 +\move, \weightedHeihgt) {\scriptsize 1};


  \node  at (14*    \step -0.15-\move, \layerHeight) {\scriptsize 7L};
  \node  at (14*   \step -0.15 -\move, \unitHeight) {\scriptsize 82U};
  \node  at (14*   \step -0.1  -\move, \weightedHeihgt) {\scriptsize 0};

  \node  at (16*    \step, \layerHeight) {\scriptsize 7L};
  \node  at (16*    \step, \unitHeight) {\scriptsize 82U};
  \node  at (16*    \step, \weightedHeihgt) {\scriptsize 0};

  \node  at (17.6*  \step +0.15+\move, \layerHeight) {\scriptsize 7L};
  \node  at (17.6*  \step +0.15+\move, \unitHeight) {\scriptsize 50U};
  \node  at (17.6*  \step +0.1 +\move, \weightedHeihgt) {\scriptsize 0};


  \node  at (24.5*  \step -0.15-\move, \layerHeight) {\scriptsize 7L};
  \node  at (24.5*  \step -0.15-\move, \unitHeight) {\scriptsize 82U};
  \node  at (24.5*  \step -0.1 -\move,  \weightedHeihgt) {\scriptsize 1};

  \node  at (26.1*  \step, \layerHeight) {\scriptsize 7L};
  \node  at (26.1*  \step, \unitHeight) {\scriptsize 82U};
  \node  at (26.1*  \step, \weightedHeihgt) {\scriptsize 0};

  \node  at (28*    \step +0.15+\move, \layerHeight) {\scriptsize 7L};
  \node  at (28*    \step +0.15+\move, \unitHeight) {\scriptsize 82U};
  \node  at (28*    \step +0.1 +\move,  \weightedHeihgt) {\scriptsize 1};


  \node  at (35*    \step -0.15-\move, \layerHeight) {\scriptsize 4L};
  \node  at (35*    \step -0.15-\move, \unitHeight) {\scriptsize 82U};
  \node  at (35*    \step -0.1 -\move,  \weightedHeihgt) {\scriptsize 0};

  \node  at (37*    \step, \layerHeight) {\scriptsize 7L};
  \node  at (37*    \step, \unitHeight) {\scriptsize 82U};
  \node  at (37*    \step, \weightedHeihgt) {\scriptsize 0};

  \node  at (38.7*  \step +0.15+\move, \layerHeight) {\scriptsize 4L};
  \node  at (38.7*  \step +0.15+\move, \unitHeight) {\scriptsize 50U};
  \node  at (38.7*  \step +0.1 +\move,  \weightedHeihgt) {\scriptsize 1};

  \begin{axis}[
    major x tick style = transparent,
    ybar=2*\pgflinewidth,
    x=10pt,
    ymajorgrids = true,
    ylabel = {Accuracy},
    ylabel style = {font=\footnotesize},
    xtickmin={1},
    xtickmax={21},
    axis x line = bottom,
    axis y line = left,
    enlarge y limits={upper, value=0.1},
    xticklabels = {,,},
    bar shift=0pt,
    enlarge x limits=0.1,
    ymin=0.76,
    ymax=0.98,
    legend style={at={(0.5, -0.4)}, anchor=north, legend columns=3},
    every axis legend/.append style={nodes={right}},
    nodes near coords={\vspace*{0.1\baselineskip}
      \foreach \X in \pgfplotspointmeta%
      {\centerline{\X}\newline}%
      \vspace*{-0.7\baselineskip}
    },
    nodes near coords style={font=\scriptsize,anchor=-90, text width=1cm},
    ]

    \addplot[pointBiLSTM=blue] coordinates {

    (1, 0.9025442765600247) 
    (7, 0.8766640656404435) 
    (13, 0.8988593838876747) 
    (19, 0.9638475586025206) 
    };

    \addplot[pointLSTM=blue] coordinates {
    (1, 0.7977761225792721) 
    (7, 0.7883237568276937) 
    (13, 0.827841810615813) 
    (19, 0.816387749603329) 
    };

    \addplot[pointBiLSTM=red] coordinates {
    (2, 0.9346846846846846) 
    (8, 0.9310432479723819) 
    (14, 0.9472562344699578) 
    (20, 0.9280962787773555) 
    };

    \addplot[pointLSTM=red] coordinates {
    (2, 0.8965678276701898) 
    (8, 0.8869854106074577) 
    (14, 0.9094811843247613) 
    (20, 0.9219171930664911) 
    };

    \addplot[pointBiLSTM=brown] coordinates {
    (3, 0.9458158946347922) 
    (9, 0.9411340474332599) 
    (15, 0.954692692273149) 
    (21, 0.9628835733317367) 
    };

    \addplot[pointLSTM=brown] coordinates {
    (3, 0.9170331749071906) 
    (9, 0.8972239956491926) 
    (15, 0.94120887345448) 
    (21, 0.9284854653773613) 
    };

    \addplot[pointRug=blue, mark size=0.7pt]
    coordinates {

%
(1, 0.861495353621338) 
(1, 0.866862925918044) 
(1, 0.867725993710246) 
(1, 0.8706225910950319) 
(1, 0.8794601688302476) 
(1, 0.8869558534912864) 
(1, 0.8948003121231469) 
(1, 0.9025442765600247) 

(1, 0.23921165259747937) 
(1, 0.23921165259747937) 
(1, 0.23921165259747937) 
(1, 0.23921165259747937) 
(1, 0.5386134165661725) 
(1, 0.6374642358894328) 
(1, 0.6666016410110899) 
(1, 0.7977761225792721) 

(7, 0.7722328627840438) 
(7, 0.7918174079591402) 
(7, 0.7977524768863351) 
(7, 0.8173251992149632) 
(7, 0.8267420964271357) 
(7, 0.8307795984961339) 
(7, 0.8564115296398761) 
(7, 0.8766640656404435) 

(7, 0.23921165259747937) 
(7, 0.23921165259747937) 
(7, 0.23921165259747937) 
(7, 0.5082878153744296) 
(7, 0.7268331323449435) 
(7, 0.7272055520087017) 
(7, 0.7346362110141638) 
(7, 0.7883237568276937) 

(13, 0.8288896206927522) 
(13, 0.8555698589947012) 
(13, 0.8634134658563603) 
(13, 0.8649043499086908) 
(13, 0.8741370535580637) 
(13, 0.8841181929766787) 
(13, 0.8986558093584409) 
(13, 0.8988593838876747) 

(13, 0.6457563691883963) 
(13, 0.740628087297548) 
(13, 0.7581235218393557) 
(13, 0.7709068047780139) 
(13, 0.7879471903721222) 
(13, 0.8079693440708917) 
(13, 0.8203873903541598) 
(13, 0.827841810615813) 

(19, 0.8182558452833578) 
(19, 0.955668652516241) 
(19, 0.9576864353501182) 
(19, 0.9604047540640063) 
(19, 0.9611891147501722) 
(19, 0.9619555129778763) 
(19, 0.9621052001317247) 
(19, 0.9638475586025206) 

(19, 0.6293147322096817) 
(19, 0.6499715594407689) 
(19, 0.686908361524414) 
(19, 0.7040744843277551) 
(19, 0.7495314792084542) 
(19, 0.7611831272640182) 
(19, 0.783803849953597) 
(19, 0.816387749603329) 
    };

    \addplot[pointRug=red, mark size=0.7pt]
    coordinates {
%
(2, 0.23921165259747937) 
(2, 0.23921165259747937) 
(2, 0.9222411387765719) 
(2, 0.9248953678087536) 
(2, 0.9309841337400391) 
(2, 0.9320954813080796) 
(2, 0.9336324513489865) 
(2, 0.9346846846846846) 

(2, 0.23921165259747937) 
(2, 0.23921165259747937) 
(2, 0.23921165259747937) 
(2, 0.23921165259747937) 
(2, 0.23921165259747937) 
(2, 0.23921165259747937) 
(2, 0.8536213378733064) 
(2, 0.8965678276701898) 

(8, 0.23921165259747937) 
(8, 0.23921165259747937) 
(8, 0.9026743278711783) 
(8, 0.9032713816178385) 
(8, 0.914609491381145) 
(8, 0.9212184625570452) 
(8, 0.9275673311106383) 
(8, 0.9310432479723819) 

(8, 0.23921165259747937) 
(8, 0.23921165259747937) 
(8, 0.23921165259747937) 
(8, 0.23921165259747937) 
(8, 0.23921165259747937) 
(8, 0.23921165259747937) 
(8, 0.23921165259747937) 
(8, 0.8869854106074577) 

(14, 0.24334939975451306) 
(14, 0.9330898422297398) 
(14, 0.9334371164266682) 
(14, 0.9352632997036194) 
(14, 0.9392928778852199) 
(14, 0.9400592761129241) 
(14, 0.9401011885160016) 
(14, 0.9472562344699578) 

(14, 0.24334939975451306) 
(14, 0.24334939975451306) 
(14, 0.24334939975451306) 
(14, 0.24334939975451306) 
(14, 0.24334939975451306) 
(14, 0.24431937251145108) 
(14, 0.8778253450288898) 
(14, 0.9094811843247613) 

(20, 0.24334939975451306) 
(20, 0.9012124659461724) 
(20, 0.9170373918510313) 
(20, 0.9248031613926893) 
(20, 0.9251264856450019) 
(20, 0.926371882765021) 
(20, 0.9275933299404245) 
(20, 0.9280962787773555) 

(20, 0.24334939975451306) 
(20, 0.24334939975451306) 
(20, 0.24334939975451306) 
(20, 0.24334939975451306) 
(20, 0.8890099691644463) 
(20, 0.9013202406969434) 
(20, 0.910774481334012) 
(20, 0.9219171930664911) 

    };
    \addplot[pointRug=brown, mark size=0.7pt]
    coordinates {

%
(3, 0.23921165259747937) 
(3, 0.9363812631529167) 
(3, 0.9417192783334514) 
(3, 0.9424582062377337) 
(3, 0.9427892459388522) 
(3, 0.9440129105483437) 
(3, 0.9446217871414722) 
(3, 0.9458158946347922) 

(3, 0.23921165259747937) 
(3, 0.23921165259747937) 
(3, 0.23921165259747937) 
(3, 0.23921165259747937) 
(3, 0.23921165259747937) 
(3, 0.23921165259747937) 
(3, 0.9124399990541722) 
(3, 0.9170331749071906) 

(9, 0.23921165259747937) 
(9, 0.9195396183585159) 
(9, 0.9307122082712634) 
(9, 0.9329644605235157) 
(9, 0.933957579626871) 
(9, 0.9357664751365539) 
(9, 0.9373034451774608) 
(9, 0.9411340474332599) 

(9, 0.23921165259747937) 
(9, 0.23921165259747937) 
(9, 0.23921165259747937) 
(9, 0.23921165259747937) 
(9, 0.23921165259747937) 
(9, 0.23921165259747937) 
(9, 0.23921165259747937) 
(9, 0.8972239956491926) 

(15, 0.9441487291560637) 
(15, 0.9455737508607011) 
(15, 0.9467353231745651) 
(15, 0.9522318354638806) 
(15, 0.9523276352423435) 
(15, 0.9530221836362003) 
(15, 0.9537646319192887) 
(15, 0.954692692273149) 

(15, 0.24334939975451306) 
(15, 0.24334939975451306) 
(15, 0.8684369667395144) 
(15, 0.8771727090381103) 
(15, 0.9143310481094512) 
(15, 0.9250725982696165) 
(15, 0.9335987785528246) 
(15, 0.94120887345448) 

(21, 0.24334939975451306) 
(21, 0.9380534682513546) 
(21, 0.9396281771098404) 
(21, 0.9427356824237344) 
(21, 0.9445379157560698) 
(21, 0.959949705116307) 
(21, 0.9627219112055804) 
(21, 0.9628835733317367) 

(21, 0.24334939975451306) 
(21, 0.24334939975451306) 
(21, 0.24334939975451306) 
(21, 0.24334939975451306) 
(21, 0.24334939975451306) 
(21, 0.9100978953986167) 
(21, 0.9255635721342395) 
(21, 0.9284854653773613) 

    };


    \addplot[pointBiLSTM=black] coordinates {(1, 0)};\label{BiLSTM}
    \addplot[pointLSTM=black]   coordinates {(1, 0)};\label{LSTM}  
    \addplot[pointRug=blue]     coordinates {(1, 0)};\label{Binary}
    \addplot[pointRug=red]      coordinates {(1, 0)};\label{OneHot}
    \addplot[pointRug=brown]    coordinates {(1, 0)};\label{TwoHot}
  \end{axis}

    \node [draw,fill=white] at (4.2,-1.4) {\shortstack[l]{
    \ref{BiLSTM} \scriptsize{BiLSTM}
    \ref{LSTM} \scriptsize{LSTM}}};

    \node [draw,fill=white] at (4.2,-2) {\shortstack[l]{
    \ref{Binary} \scriptsize{\scriptsize{Binary(BinE)}}
    \ref{OneHot} \scriptsize{\scriptsize{One-Hot(OneE)}}
    \ref{TwoHot} \scriptsize{\scriptsize{Two-Hot(TwoE)}}
    }};

\end{tikzpicture}


%% file: results_ar_per_class_accuracy.tex
\begin{tikzpicture}[scale=1]
  \begin{axis}[
    axis x line = bottom,
    axis y line = left,
    ymajorgrids = true,
    ybar,
    enlargelimits=0.15,
    legend style={at={(0.87,0.99)},
      anchor=north,legend columns=1},
    ylabel={Accuracy},
    ylabel style = {font=\footnotesize},
    symbolic x coords={Taweel, Kamel, Baseet, Khafeef, Wafeer, Rigz, Raml, Motakarib, Sar'e, Monsareh, Mogtath, Madeed, Hazg, Motadarik, Moktadib, Modar'e},
    xtick=data,
    xticklabel style = {font=\footnotesize},
    nodes near coords align={vertical},
    x tick label style={rotate=90, anchor=east},
    bar width=2pt,
    ymin=0.6,
    ]

    \addplot[mark=triangle, every mark/.append style={rotate=180},
    thin, only marks, mark size=1.5pt, point meta=explicit symbolic, opacity=0.4]
    coordinates {
      (Wafeer,    0.9811213222198475)       
      (Monsareh,  0.9045746962115797)       
      (Mogtath,   0.897708216880939 )       
      (Motakarib, 0.9609120521172638)       
      (Kamel,     0.966898378020523 )       
      (Taweel,    0.9844991757498216)       
      (Sar'e,     0.9066397034041119)       
      (Raml,      0.9485771342985522)       
      (Rigz,      0.8889925373134329)       
      (Khafeef,   0.9661488673139158)       
      (Baseet,    0.9814341393906374) 
      (Madeed,     0)  
      (Hazg,       0) (Motadarik,  0) 
      (Moktadib,   0) 
      (Modar'e,    0)
    };

    \addplot[mark=o, thin, only marks, mark size=1.5pt, point
    meta=explicit symbolic, opacity=0.4]
    coordinates{
      (Wafeer,     0.9009759920210871)       
      (Monsareh,   0.6208005718370264)       
      (Mogtath,    0.6880939072107323)       
      (Motakarib,  0.8506281991624011)       
      (Kamel,      0.8910956636875207)       
      (Taweel,     0.9508402430922914)       
      (Sar'e,      0.6083586113919784)       
      (Raml,       0.7822016974538193)       
      (Rigz,       0.6512042062415196)       
      (Khafeef,    0.8456957928802589)       
      (Baseet,     0.9128703742508696) 
    };

    \addplot[mark=triangle*, every mark/.append style={rotate=180},
    thin, only marks, mark size=1.5pt, point meta=explicit symbolic, opacity=0.4]
    coordinates {
      (Wafeer,    0.889584964761159  )       
      (Monsareh,  0.6716101694915254 )       
      (Madeed,    0.45979899497487436)  
      (Mogtath,   0.6439135381114903 )       
      (Motakarib, 0.8148088917319687 )       
      (Kamel,     0.8776972469479428 )       
      (Taweel,    0.9439529481540059 )       
      (Sar'e,     0.6370179948586119 )       
      (Raml,      0.7719209325899645 )       
      (Rigz,      0.64107517849643   )       
      (Khafeef,   0.8741908607319105 )       
      (Baseet,    0.9208657001620072 ) 
      (Moktadib,  0.16326530612244897) 
      (Hazg,      0.5506493506493506 )
      (Modar'e,   0.0                )
      (Motadarik, 0.28               ) 
    };

    \addplot[mark=*, thin, only marks, mark size=1.5pt, point
    meta=explicit symbolic, opacity=0.4]
    coordinates {
      (Wafeer,    0.9014024346835623 )       
      (Monsareh,  0.7305790960451978 )       
      (Madeed,    0.5062814070351759 )  
      (Mogtath,   0.6814562002275313 )       
      (Motakarib, 0.8488725411802335 )       
      (Kamel,     0.9185383195083638 )       
      (Taweel,    0.9563337122522612 )       
      (Sar'e,     0.7830334190231363 )       
      (Raml,      0.8205778003041054 )       
      (Rigz,      0.6881982360352793 )       
      (Khafeef,   0.8859834647183235 )       
      (Baseet,    0.9370664229634861 ) 
      (Moktadib,  0.3673469387755102 ) 
      (Hazg,      0.6519480519480519 )
      (Modar'e,   0.08333333333333333)
      (Motadarik, 0.40190476190476193) 
    };

    \legend{
      {\scriptsize (1T, 0D)},
      {\scriptsize (1T, 1D)},
      {\scriptsize (0T, 0D)},
      {\scriptsize (0T, 1D)},
    }

    \draw[loosely dotted] (axis cs:Wafeer,    0) -- (axis cs:Wafeer,     0.789584964761159);
    \draw[loosely dotted] (axis cs:Monsareh,  0) -- (axis cs:Monsareh,   0.5208005718370264);
    \draw[loosely dotted] (axis cs:Mogtath,   0) -- (axis cs:Mogtath,    0);
    \draw[loosely dotted] (axis cs:Motakarib, 0) -- (axis cs:Motakarib,  0.7148088917319687);
    \draw[loosely dotted] (axis cs:Kamel,     0) -- (axis cs:Kamel,      0.7776972469479428);
    \draw[loosely dotted] (axis cs:Taweel,    0) -- (axis cs:Taweel,     0.8439529481540059);
    \draw[loosely dotted] (axis cs:Sar'e,     0) -- (axis cs:Sar'e,      0.5083586113919784);
    \draw[loosely dotted] (axis cs:Raml,      0) -- (axis cs:Raml,       0.6719209325899645);
    \draw[loosely dotted] (axis cs:Rigz,      0) -- (axis cs:Rigz,       0.54107517849643);
    \draw[loosely dotted] (axis cs:Khafeef,   0) -- (axis cs:Khafeef,    0.7456957928802589);
    \draw[loosely dotted] (axis cs:Baseet,    0) -- (axis cs:Baseet,     0.8128703742508696);
    \draw[loosely dotted] (axis cs:Moktadib,  0) -- (axis cs:Moktadib,   0);
    \draw[loosely dotted] (axis cs:Madeed,    0) -- (axis cs:Madeed,     0.35979899497487436);
    \draw[loosely dotted] (axis cs:Hazg,      0) -- (axis cs:Hazg,       0.4506493506493506);
    \draw[loosely dotted] (axis cs:Modar'e,   0) -- (axis cs:Modar'e,    0);
    \draw[loosely dotted] (axis cs:Motadarik, 0) -- (axis cs:Motadarik,  0.18);


    \draw[line width=0.1pt, densely dashed]
    (axis cs:Taweel,        0.9844991757498216)
    -- (axis cs:Kamel,      0.966898378020523 )
    -- (axis cs:Baseet,     0.9814341393906374)
    -- (axis cs:Khafeef,    0.9661488673139158)
    -- (axis cs:Wafeer,     0.9811213222198475)
    -- (axis cs:Rigz,       0.8889925373134329)
    -- (axis cs:Raml,       0.9485771342985522)
    -- (axis cs:Motakarib,  0.9609120521172638)
    -- (axis cs:Sar'e,      0.9066397034041119)
    -- (axis cs:Monsareh,   0.9045746962115797)
    -- (axis cs:Mogtath,    0.897708216880939 );

   \draw[line width=0.1pt, densely dashed]
   (axis cs:Taweel,        0.9508402430922914)
   -- (axis cs:Kamel,      0.8910956636875207)
   -- (axis cs:Baseet,     0.9128703742508696)
   -- (axis cs:Khafeef,    0.8456957928802589)
   -- (axis cs:Wafeer,     0.9009759920210871)
   -- (axis cs:Rigz,       0.6512042062415196)
   -- (axis cs:Raml,       0.7822016974538193)
   -- (axis cs:Motakarib,  0.8506281991624011)
   -- (axis cs:Sar'e,      0.6083586113919784)
   -- (axis cs:Monsareh,   0.6208005718370264)
   -- (axis cs:Mogtath,    0.6880939072107323);

    \draw[line width=0.1pt, densely dashed]
    (axis cs:Taweel,        0.9439529481540059 )
    -- (axis cs:Kamel,      0.8776972469479428 )
    -- (axis cs:Baseet,     0.9208657001620072 )
    -- (axis cs:Khafeef,    0.8741908607319105 )
    -- (axis cs:Wafeer,     0.889584964761159  )
    -- (axis cs:Rigz,       0.64107517849643   )
    -- (axis cs:Raml,       0.7719209325899645 )
    -- (axis cs:Motakarib,  0.8148088917319687 )
    -- (axis cs:Sar'e,      0.6370179948586119 )
    -- (axis cs:Monsareh,   0.6716101694915254 )
    -- (axis cs:Mogtath,    0.6439135381114903 )
    -- (axis cs:Madeed,     0.45979899497487436)
    -- (axis cs:Hazg,       0.5506493506493506 )
    -- (axis cs:Motadarik,  0.28               )
    -- (axis cs:Moktadib,   0.16326530612244897)
    -- (axis cs:Modar'e,    0.0                );

   \draw[line width=0.1pt, densely dashed]
   (axis cs:Taweel,        0.9563337122522612 )
   -- (axis cs:Kamel,      0.9185383195083638 )
   -- (axis cs:Baseet,     0.9370664229634861 )
   -- (axis cs:Khafeef,    0.8859834647183235 )
   -- (axis cs:Wafeer,     0.9014024346835623 )
   -- (axis cs:Rigz,       0.6881982360352793 )
   -- (axis cs:Raml,       0.8205778003041054 )
   -- (axis cs:Motakarib,  0.8488725411802335 )
   -- (axis cs:Sar'e,      0.7830334190231363 )
   -- (axis cs:Monsareh,   0.7305790960451978 )
   -- (axis cs:Mogtath,    0.6814562002275313 )
   -- (axis cs:Madeed,     0.5062814070351759 )
   -- (axis cs:Hazg,       0.6519480519480519 )
   -- (axis cs:Motadarik,  0.40190476190476193)
   -- (axis cs:Moktadib,   0.3673469387755102 )
   -- (axis cs:Modar'e,    0.08333333333333333);

  \end{axis}
\end{tikzpicture}


%% file: results_ar_convergence.tex
\begin{axis}[
  height=7cm,
  grid=major,
  every axis y label/.style= {at={( 0.1, 1.1)}, anchor=north},
  xlabel={Epoch},
  ylabel={Accuracy},
  legend style={at={(0.6, 0.18)},anchor=west},
  every axis legend/.append style={nodes={right}},
  ymin = 0.85,
  name=left plot,
  title style={at={(0.5,-.4)}}
  ]

  \addplot[color=brown, mark=*, thick, mark size=0.5pt] coordinates {

    (1,      0.4973408579826355 )
    (2,      0.6637325882911682 )
    (3,      0.7843776345252991 )
    (4,      0.8562803268432617 )
    (5,      0.8955804109573364 )
    (6,      0.9072529673576355 )
    (7,      0.9148279428482056 )
    (8,      0.9249432682991028 )
    (9,      0.9284470081329346 )
    (10,     0.9312249422073364 )
    (11,     0.9338576793670654 )
    (12,     0.9358240365982056 )
    (13,     0.9362859129905701 )
    (14,     0.9381598234176636 )
    (15,     0.9399744272232056 )
    (16,     0.9379156827926636 )
    (17,     0.9388328790664673 )
    (18,     0.9388262629508972 )
    (19,     0.939400315284729 )
    (20,     0.9404956698417664 )
    (21,     0.9401789307594299 )
    (22,     0.9387009143829346 )
    (23,     0.9391694068908691 )
    (24,     0.9394267201423645 )
    (25,     0.939261794090271 )
    (26,     0.9391694068908691 )
    (27,     0.9397236704826355 )
    (28,     0.9380608797073364 )
    (29,     0.9386216998100281 )
    (30,     0.9390044212341309 )
    (31,  0.9392420053482056)
    (32,  0.9390110373497009)
    (33,  0.9363254904747009)
    (34,  0.9365366101264954)
    (35,  0.9395982623100281)
    (36,  0.9375395774841309)
    (37,  0.9366289973258972)
    (38,  0.9384303689002991)
    (39,  0.9370908737182617)
    (40,  0.9396642446517944)
  };

  \addplot[color=red, mark=*, thick, mark size=0.5pt] coordinates {
    (1,        0.5241501331329346 )
    (2,        0.6912611126899719 )
    (3,        0.8391179442405701 )
    (4,        0.8904072642326355 )
    (5,        0.9107500910758972 )
    (6,        0.9256294965744019 )
    (7,        0.9318781495094299 )
    (8,        0.9354742765426636 )
    (9,        0.9377507567405701 )
    (10,       0.9410961270332336 )
    (11,       0.9423761963844299 )
    (12,       0.9436959028244019 )
    (13,       0.9424092173576355 )
    (14,       0.9431548118591309 )
    (15,       0.9411951303482056 )
    (16,       0.9415910243988037 )
    (17,       0.942330002784729 )
    (18,       0.9428513050079346 )
    (19,       0.9439400434494019 )
    (20,       0.9432538151741028 )
    (21,       0.9424421787261963 )
    (22,       0.9430954456329346 )
    (23,       0.9435111284255981 )
    (24,       0.9440720081329346 )
    (25,       0.9439598321914673 )
    (26,       0.9431614279747009 )
    (27,       0.9429964423179626 )
    (28,       0.9439268112182617 )
    (29,       0.9410235285758972 )
    (30,       0.9415580034255981 )
    (31,  0.9408321976661682)
    (32,  0.9407595992088318)
    (33,  0.9425477981567383)
    (34,  0.939453125 )
    (35,  0.9388592839241028)
    (36,  0.942191481590271 )
    (37,  0.9434715509414673)
    (38,  0.9433857798576355)
    (39,  0.9416041970252991)
    (40,  0.9427457451820374)
  };

  \addplot[color=blue, mark=*, thick, mark size=0.5pt] coordinates {
    (1,         0.4176618158817291)
    (2,         0.7385522127151489)
    (3,         0.8620811104774475)
    (4,         0.8990639448165894)
    (5,         0.9152568578720093)
    (6,         0.9270322322845459)
    (7,         0.9314031600952148)
    (8,         0.9404243230819702)
    (9,         0.9433448910713196)
    (10,        0.9482346773147583)
    (11,        0.9494454860687256)
    (12,        0.9520400762557983)
    (13,        0.9542421698570251)
    (14,        0.956171452999115 )
    (15,        0.9566637873649597)
    (16,        0.9575485587120056)
    (17,        0.9581539630889893)
    (18,        0.9601032733917236)
    (19,        0.9615402817726135)
    (20,        0.9603028297424316)
    (21,        0.9616068005561829)
    (22,        0.9614471197128296)
    (23,        0.9629971981048584)
    (24,        0.9632699489593506)
    (25,        0.9633697867393494)
    (26,        0.9635560512542725)
    (27,        0.9630637168884277)
    (28,        0.9634895324707031)
    (29,        0.9636757969856262)
    (30,        0.9642080068588257)
    (31, 0.963502824306488 )
    (32, 0.9632300734519958)
    (33, 0.9634096622467041)
    (34, 0.9632500410079956)
    (35, 0.963416337966919 )
    (36, 0.9633830785751343)
    (37, 0.962651252746582 )
    (38, 0.962751030921936 )
    (39, 0.9625581502914429)
    (40, 0.9630969762802124)
  };

  \legend{
    {\scriptsize Two-Hot},
    {\scriptsize One-Hot},
    {\scriptsize Binary},
  }

\end{axis}


%% file: results_en_models_acc.tex
\begin{axis}[
  major x tick style = transparent,
  height=7cm,
  legend style={at={(0.7, 0.2)},anchor=west},
  every axis legend/.append style={nodes={right}},
  name=left plot,
  xtick distance=1,
  axis x line = bottom,
  axis y line = left,
  xmin=0,
  xmax=3.5,
  x=1.5cm,
  title={\small (a) \textit{Overall accuracy}},
  title style={at={(0.5,-.4)}},
  xticklabels = {, ,},
  ymajorgrids = true,
  nodes near coords,
  ylabel = {Accuracy},
  ylabel style = {font=\footnotesize},
  ymin=0.72,
  ymax=0.857,
  ]

  \addplot[pointBiGRU=blue] coordinates {
(1, 0.8169014084507042) 

  };
  \addplot[pointGRU=blue] coordinates {
(1, 0.7859154929577463) 
  };
  \addplot[pointLSTM=blue] coordinates {
(1, 0.7690140845070422) 

  };
  \addplot[pointBiLSTM=blue] coordinates {
(1, 0.7661971830985915) 

  };

  \addplot[pointBiGRU=red] coordinates {
    (2, 0.8264788732394366) 

  };
  \addplot[pointGRU=red] coordinates {
(2, 0.8214084507042254) 

  };
  \addplot[pointLSTM=red] coordinates {
(2, 0.8022535211267606) 

  };
  \addplot[pointBiLSTM=red] coordinates {
(2, 0.7650704225352113) 

  };

  \addplot[pointModelsFiguresStyle=blue, mark size=0.7pt, mark=-]
  coordinates {
(1, 0.7138028169014086) 
(1, 0.7211267605633802) 
(1, 0.7245070422535211) 
(1, 0.7256338028169014) 
(1, 0.7278873239436621) 
(1, 0.7312676056338029) 
(1, 0.7357746478873239) 
(1, 0.7380281690140845) 
(1, 0.7414084507042253) 
(1, 0.7425352112676056) 
(1, 0.7447887323943663) 
(1, 0.7447887323943663) 
(1, 0.7498591549295774) 
(1, 0.7526760563380281) 
(1, 0.7543661971830986) 
(1, 0.7571830985915493) 
(1, 0.7588732394366198) 
(1, 0.7622535211267607) 
(1, 0.7650704225352112) 
(1, 0.7661971830985915) 
(1, 0.7661971830985916) 
(1, 0.7684507042253521) 
(1, 0.7690140845070422) 
(1, 0.7735211267605634) 
(1, 0.7752112676056337) 
(1, 0.7780281690140846) 
(1, 0.7808450704225353) 
(1, 0.7825352112676058) 
(1, 0.7859154929577463) 
(1, 0.7870422535211268) 
(1, 0.7954929577464789) 
(1, 0.7988732394366197) 
(1, 0.8011267605633803) 
(1, 0.8022535211267606) 
(1, 0.8050704225352113) 
(1, 0.8169014084507042) 
  };
  \addplot[pointModelsFiguresStyle=red, mark size=0.7pt, mark=-]
  coordinates {
(2, 0.7059154929577465) 
(2, 0.716056338028169) 
(2, 0.7295774647887323) 
(2, 0.7307042253521127) 
(2, 0.7329577464788732) 
(2, 0.7340845070422535) 
(2, 0.7374647887323944) 
(2, 0.7380281690140844) 
(2, 0.7459154929577465) 
(2, 0.755492957746479) 
(2, 0.7583098591549295) 
(2, 0.7594366197183099) 
(2, 0.7616901408450705) 
(2, 0.7622535211267606) 
(2, 0.7628169014084507) 
(2, 0.763943661971831) 
(2, 0.7650704225352113) 
(2, 0.7684507042253521) 
(2, 0.779718309859155) 
(2, 0.7808450704225353) 
(2, 0.7836619718309861) 
(2, 0.7904225352112677) 
(2, 0.8005633802816902) 
(2, 0.8011267605633803) 
(2, 0.8022535211267606) 
(2, 0.8045070422535211) 
(2, 0.8090140845070422) 
(2, 0.8107042253521127) 
(2, 0.811830985915493) 
(2, 0.8129577464788732) 
(2, 0.815774647887324) 
(2, 0.816338028169014) 
(2, 0.8174647887323944) 
(2, 0.8180281690140845) 
(2, 0.8214084507042254) 
(2, 0.8264788732394366) 
  };

    \addplot[pointBiLSTM=black] coordinates {(1, 0)};\label{BiLSTM}
    \addplot[pointLSTM=black]   coordinates {(1, 0)};\label{LSTM}
    \addplot[pointBiGRU=black]  coordinates {(1, 0)};\label{BiGRU}
    \addplot[pointGRU=black]    coordinates {(1, 0)};\label{GRU}
    \addplot[pointRug=red]      coordinates {(1, 0)};\label{OneHot}
    \addplot[pointRug=blue]     coordinates {(1, 0)};\label{Binary}
\end{axis}

    \node [draw,fill=white] at (4.3,1.2) {\shortstack[l]{
    \ref{BiGRU} \scriptsize{BiGRU}\\
    \ref{GRU} \scriptsize{GRU}\\
    \ref{LSTM} \scriptsize{LSTM}\\
    \ref{BiLSTM} \scriptsize{BiLSTM}
    }};

    \node [draw,fill=white] at (2.6,-0.9) {\shortstack[l]{
    \ref{Binary} \scriptsize{\scriptsize{Binary(BinE)}}
    \ref{OneHot} \scriptsize{\scriptsize{One-Hot(OneE)}}
    }};

\node at (7.5 * 0.2, 5.2) {\scriptsize 7L};
\node at (7.5 * 0.2, 5.2 -0.3) {\scriptsize 40U};

\node at (15 * 0.2, 5.2) {\scriptsize 4L};
\node at (15 * 0.2, 5.2 -0.3) {\scriptsize 40U};


%% file: results_en_per_class_acc.tex
\begin{axis}[
  symbolic x coords={Iambic,Trochaic,Anapaestic,Dactyl},
  xtick=data,
  every axis y label/.style= {at={( 0.1, 1.1)}, anchor=north},
  x tick label style={font=\footnotesize},
  height=7cm,
  x=0.5cm,
  ylabel={Accuracy},
  enlarge x limits=0.3,
  enlarge y limits=0.07,
  x tick label style={rotate=90, anchor=east},
  /pgf/bar width=20pt,
  at={($(left plot.south east)+(1.5cm,0)$)},
  title={\small (b) \textit{Per-class Accuracy}},
  title style={at={(0.5,-.4)}}
  ]

    \addplot[mark=triangle*, every mark/.append style={rotate=180},
thin, only marks, mark size=1.5pt, point meta=explicit symbolic, opacity=0.4]
   coordinates {
      (Iambic,     0.9099264705882353)
      (Anapaestic, 0.9788732394366197)
      (Trochaic,   0.8827977315689981)
      (Dactyl,     0.6535714285714286)
    };
    

  \draw[line width=0.1pt, densely dashed]
      (axis cs:Iambic,       0.9099264705882353)
   -- (axis cs:Trochaic,     0.8827977315689981 )
   -- (axis cs:Anapaestic,   0.9788732394366197)
   -- (axis cs:Dactyl,       0.6535714285714286);

\end{axis}
